\documentclass[10pt,twocolumn,letterpaper]{article}

\usepackage[pagenumbers]{cvpr} 

\usepackage{newtxtext,newtxmath}
\usepackage{epsfig}
\usepackage{graphicx}
\usepackage{amsmath}
\usepackage{amssymb}
\usepackage{amsfonts}
\usepackage{booktabs}
\usepackage{multirow}
\usepackage{array}
\usepackage{makecell}
\usepackage{xcolor}
\usepackage{colortbl}
\definecolor{gainGreen}{HTML}{3C7A3C}
\definecolor{lossRed}{HTML}{A33333}

\newcommand{\gain}[1]{\textcolor{gainGreen}{\scriptsize(+#1)}}
\newcommand{\loss}[1]{\textcolor{lossRed}{\scriptsize($-$#1)}}
\newlength{\cnumwd}
\newlength{\cstarwd}
\newlength{\cannwd}
\newcommand{\cwin}[2]{\makebox[\cnumwd][r]{\textbf{\underline{#1}}}\makebox[\cstarwd][c]{\textcolor{lossRed}{\scriptsize$\bigstar$}}\makebox[\cannwd][l]{#2}}
\newcommand{\clos}[2]{\makebox[\cnumwd][r]{#1}\makebox[\cstarwd][c]{}\makebox[\cannwd][l]{#2}}
\newcommand{\cneu}[1]{\makebox[\cnumwd][r]{\textbf{#1}}\makebox[\cstarwd][c]{}\makebox[\cannwd][l]{}}
\newcommand{\oc}[1]{\cellcolor{black!7}#1}
\newcommand{\rsurv}[1]{\cellcolor{green!16}#1}
\newcommand{\rbrk}[1]{\cellcolor{red!12}#1}
\usepackage{placeins}
\usepackage{algorithm}
\usepackage{algpseudocode}
\usepackage[pagebackref,breaklinks,colorlinks]{hyperref}

\newcommand{\method}{\textsc{Chord}}
\newcommand{\num}[1]{\textup{#1}}
\newcommand{\best}[1]{\textbf{#1}}
\newcommand{\good}{\textcolor{black!60!green}{$\checkmark$}}

\newcommand{\bad}{\textcolor{red!70!black}{$\times$}}
\newcommand{\lpc}{\ensuremath{\mathcal{L}_{\mathrm{pc}}}}
\newcommand{\lgf}{\ensuremath{\mathcal{L}_{\mathrm{gf}}}}
\newcommand{\lent}{\ensuremath{\mathcal{L}_{\mathrm{ent}}}}
\newcommand{\ldet}{\ensuremath{\mathcal{L}_{\mathrm{det}}}}

\newcommand{\FPR}{\mathrm{FPR}}

\def\confName{CVPR}
\def\confYear{2027}

\begin{document}

\title{\method: Learned Chunk-Conditioned Watermarking\\ 
    for Autoregressive Image Generation}

\author{
Zheyu Fu\thanks{Equal contribution. fuzy23@mails.tsinghua.edu.cn}\\
Tsinghua University\\
\and
Leyi Pan\footnotemark[1]\\
Tsinghua University\\
\and
Ruchang Yao\\
MMLab, CUHK\\
\and
Lijie Wen\thanks{Corresponding author. wenlj@mail.tsinghua.edu.cn}\\
Tsinghua University\\
}

\maketitle

\begin{abstract}
Autoregressive (AR) image generators sample discrete vector-quantized (VQ) tokens, where watermarks bias logits toward a \emph{greenlist} and verify images by scoring re-encoded tokens. Yet content-preserving edits alter 10\% to 96\% of recovered tokens. This token drift creates a dilemma: hashing preceding token IDs is brittle as any change randomizes downstream greenlists, whereas a fixed global greenlist survives edits but leaves frequency biases that let attackers infer and remove it. How can greenlists vary with context for secrecy yet remain recoverable after edits?

To resolve this dilemma, we propose \method{}, summarizing each preceding token chunk by its normalized mean codebook embedding rather than discrete IDs. A lightweight network smoothly maps this summary to vocabulary biases while decorrelating greenlists across contexts. Since edits mostly replace tokens with nearby codebook neighbors, recomputed greenlists stay stable after editing yet vary across images.

On LlamaGen-B, \method{} keeps high post-edit greenlist overlap while exact chunk hashing drops to chance, matching global KGW on aligned edits and beating one-token KGW by 0.175 mean AUROC on rotation and cropping. With 500 queried images, frequency removal collapses global KGW to 0.015 AUROC, whereas \method{} retains 0.943 and holds context-clustering inference near chance (0.536 accuracy). \method{} thus resists both edits and rule extraction while surpassing existing AR watermarks on diffusion regeneration and most geometric attacks across LlamaGen-B and LlamaGen-L at comparable quality. \emph{Code: https://github.com/THU-BPM/CHORD}
\end{abstract}
\section{Introduction}
\label{sec:intro}

\begin{figure}[t]
    \centering
    \includegraphics[width=\columnwidth]{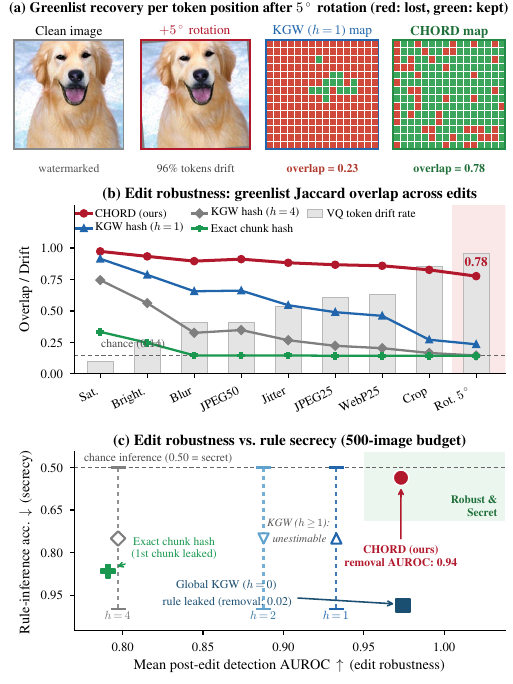}
    \caption{\textbf{Edit robustness and rule secrecy in AR image watermarking.}
    \textbf{(a)}~Rotating a watermarked LlamaGen-B image by $5^\circ$ alters
    \num{96}\% of VQ tokens: one-token hashed KGW ($h{=}1$) loses its
    greenlists across positions (\num{0.23} mean overlap), whereas \method{}
    preserves them (\num{0.78}).
    \textbf{(b)}~Across nine edits (\num{400} images, $\gamma{=}0.25$),
    hashed baselines degrade under token drift while \method{} retains high
    greenlist overlap (\num{0.78} under rotation).
    \textbf{(c)}~Plotting post-edit detection AUROC against rule-inference
    accuracy (\num{500} queried images) shows that global KGW ($h{=}0$)
    leaks its static rule, whereas \method{} achieves both edit robustness
    and secrecy.}
    \label{fig:teaser}
\end{figure}

Diffusion models have driven photorealistic image
synthesis~\cite{NEURIPS2020_4c5bcfec,dhariwal2021diffusion,rombach2022high,peebles2023scalable},
and most watermarking tools were built around their continuous latent
spaces~\cite{wen2023treering,fernandez2023stablesignature,yang2024gaussianshading,pan2026markdiffusion}.
Meanwhile, autoregressive (AR) image generators such as
MaskGIT~\cite{chang2022maskgit}, Parti~\cite{yu2022parti},
Chameleon~\cite{team2024chameleon}, and LlamaGen~\cite{sun2024llamagen}
have closed the quality gap~\cite{Esser_2021_CVPR} by encoding images into
discrete vector-quantized (VQ) tokens and modeling sequences with a
transformer. Because generation samples discrete codebook indices rather
than Gaussian latents, diffusion watermarks do not transfer. As AR
generators move toward deployment, a central provenance question arises:
given a single image, how can we verify whether it came from a protected
AR model~\cite{pan2026omnisafetybenchbenchmarksafetyevaluation}?

Following language model watermarking~\cite{kirchenbauer2023kgw,kirchenbauer2024reliability,pan2024markllm,liu2024survey},
in-generation AR image watermarks partition the VQ codebook into green and
red sets during sampling, bias logits toward green tokens, and verify
images by measuring the green fraction of re-encoded tokens. In practice,
however, shared images are routinely compressed, cropped, rotated, or
color-adjusted. Although such edits preserve semantics, re-encoding causes
severe \emph{token drift}~\cite{müller2026robustnesswatermarkingautoregressiveimage}:
on LlamaGen-B, WebP ($q{=}25$) alters \num{63}\% of token IDs and a
$5^\circ$ rotation alters \num{96}\% (\cref{fig:teaser}b), showing that
visual preservation does not preserve discrete tokens. Existing AR
watermarks counter token drift along two lines, each with clear
limitations. At the tokenizer level, WMAR~\cite{jovanovic2025wmar}
fine-tunes the VQ autoencoder for reverse cycle consistency and adds
geometric synchronization, but retraining every distinct VQ tokenizer is
costly and prevents plug-and-play deployment across off-the-shelf
generators. At the codebook level with a frozen tokenizer,
ClusterMark~\cite{lukovnikov2026clustermark} hashes the discrete cluster
label of the preceding token, while IndexMark~\cite{tong2025indexmark},
UniMark~\cite{yilmaz2026unimark}, and BitMark~\cite{kerner2025bitmark}
rely on context-free partitions via pairwise matching, semantic grouping,
or bit-level rules. Yet ClusterMark's hard quantization into discrete
clusters remains a step function where any token drift across a cluster
boundary rewrites the greenlist entirely, whereas context-free designs
such as IndexMark go even further by reusing one static split across
images and exposing exploitable token-frequency biases.

This limitation reflects a deeper dilemma between edit robustness and rule
secrecy under a frozen VQ tokenizer. An unhashed, fixed global greenlist
($h{=}0$)~\cite{zhao2024provable,hou2024semstamp,fairoze2024publicly,liu2024semantic} avoids context mismatch and survives
aligned edits, but favoring the same tokens across images creates an
exploitable frequency bias: from \num{500} watermarked images, an attacker
infers the partition with \num{0.99} accuracy and removes it via codebook
replacement, dropping AUROC to \num{0.015} (\cref{sec:removal}). Static
splits thus sacrifice secrecy for robustness. Conversely, hashing the
preceding $h$ token IDs hides token frequencies~\cite{kirchenbauer2023kgw,aaronson2023gumbel,kuditipudi2024robust,christ2024undetectable}
but breaks under token drift: one altered context token randomizes the
greenlist even when the scored token is unchanged, and larger $h$ cascades
this mismatch downstream. Under JPEG ($q{=}25$), greenlist Jaccard overlap
falls to \num{0.49} ($h{=}1$) and \num{0.22} ($h{=}4$), nearing the
\num{0.14} chance floor (\cref{fig:teaser}b) and confirming the fragility
of exact prefix hashing. Frozen-tokenizer AR watermarking thus requires
\emph{a greenlist that varies with visual context for secrecy while
evolving smoothly under content-preserving edits.}

To meet both requirements, we propose \method{} (\textbf{CH}unk-c\textbf{O}nditioned \textbf{R}obust and \textbf{D}ecorrelated watermarking), which conditions the
watermark on the normalized mean codebook embedding of the preceding
$C{=}32$ token chunk rather than discrete IDs. A lightweight network
smoothly maps this continuous summary to vocabulary biases while
decorrelating greenlists across contexts. Because edits mostly replace
tokens with nearby codebook neighbors, pre- and post-edit chunk summaries
retain \num{0.88} to \num{0.99} cosine similarity versus \num{0.16} across
unrelated images (\cref{sec:main}), preserving edit invariance without
collapsing across images. Greenlists thus keep Jaccard overlaps of
\num{0.87} under JPEG ($q{=}25$) and \num{0.78} under $5^\circ$ rotation
(\cref{fig:teaser}b, \cref{tab:consistency}), whereas exact hashing of the
same chunks drops to chance across nearly all edits, proving that
stability comes from our learned smooth mapping rather than chunking alone.

This stability yields reliable post-edit detection without sacrificing
secrecy. On LlamaGen-B, \method{} achieves \num{0.993} mean AUROC on
aligned edits, matching global KGW ($h{=}0$), and beats one-token KGW
($h{=}1$) by \num{+0.193} on rotation and \num{+0.158} on cropping
(\cref{tab:edit_detection}). With \num{500} queried images, it retains
\num{0.943} AUROC after frequency-based removal (versus \num{0.015} for
$h{=}0$) and holds context-clustering rule inference near chance
(\num{0.536} accuracy at full coverage, versus \num{0.986} on $h{=}0$;
\cref{fig:teaser}c).
Across LlamaGen-B and LlamaGen-L, \method{} also outperforms published AR
watermarks on diffusion regeneration and most geometric attacks while
preserving visual quality, and transfers across resolutions and keys
without retraining.

Our main contributions are threefold:
(i) we reveal the tension between post-edit greenlist recovery and rule
secrecy in frozen-tokenizer AR watermarking, showing why static splits
fail frequency inference and exact hashes fail under token drift;
(ii) we propose \method{}, a plug-and-play watermark smoothly mapping
pooled chunk embeddings to decorrelated biases, transferring across
scales, resolutions, and keys without tokenizer retraining; and
(iii) across semantic-preserving edits, regeneration, and a new
context-clustering rule-inference attack for continuous conditions, we
show that \method{} uniquely withstands both edits and rule extraction.

\section{Related Work}
\label{sec:related}

\paragraph{Autoregressive image generation.}
Modern AR image generators combine a VQ tokenizer with a causal or masked
transformer~\cite{oord2016pixelrnn,salimans2017pixelcnnpp,Esser_2021_CVPR}.
The tokenizer maps an image into a grid of $N$ discrete tokens
$z=(z_1,\dots,z_N)$ by quantizing each spatial feature to its nearest
entry in a learned codebook $\mathcal{C}=\{e_1,\dots,e_V\}\subset
\mathbb{R}^{d}$ ($z_t\in\{1,\dots,V\}$), and a decoder reconstructs
pixels from this token grid. Conditioned on a class label or text prompt
$c$, a transformer models the token sequence as:
\begin{equation}
\label{eq:ar}
\begin{gathered}
p_\psi(z\mid c)=\prod_{t=1}^{N}p_\psi(z_t\mid z_{<t},c),\\[2pt]
p_\psi(\cdot\mid z_{<t},c)=\mathrm{softmax}(\ell_t),
\end{gathered}
\end{equation}
where $\ell_t\in\mathbb{R}^{V}$ denotes the next-token logits. Unlike
diffusion models operating in continuous Gaussian latent
spaces~\cite{NEURIPS2020_4c5bcfec}, AR generation consists of sequential
$V$-way categorical draws over a discrete codebook, and in-generation
watermarks embed signals directly through $\ell_t$ (\cref{sec:method}).
Existing models vary mainly in decoding order and scale:
MaskGIT~\cite{chang2022maskgit} adopts masked non-causal decoding,
Parti~\cite{yu2022parti} and Chameleon~\cite{team2024chameleon} scale to
billion-parameter text-conditioned models, and
LlamaGen~\cite{sun2024llamagen} demonstrates that a standard
LLaMA decoder~\cite{touvron2023llamaopenefficientfoundation} matches
diffusion quality on ImageNet. We evaluate primarily on LlamaGen at
$111$M and $343$M parameters due to its open-source tokenizer and
checkpoints, representative codebook size ($V{=}16\,384$), and sequence
length ($N{=}256$).

\paragraph{Watermarking generated images.}
Pre-generation diffusion watermarks embed signals into continuous latent
noise and therefore do not apply to discrete-token AR
generators~\cite{fernandez2023stablesignature,fernandez2022ssl,zhao2023recipe};
examples include Tree-Ring~\cite{wen2023treering}, which modifies the
Fourier spectrum of initial noise, and
Gaussian-Shading~\cite{yang2024gaussianshading}, which samples initial
latents from a key-conditioned code. Post-hoc pixel-domain watermarks,
including StegaStamp~\cite{tancik2020stegastamp}, HiDDeN-style encoders,
and the DWT-DCT variant~\cite{khan2013digitalwatermarkingimageauthenticationbased}
shipped with Stable Diffusion, embed bitstrings directly into output
pixels regardless of the generator, but are vulnerable to regeneration
attacks~\cite{saberi2023robustness,kim2020webtoon}. In a preliminary test on $100$
paired LlamaGen-B images, DWT-DCT reaches only $0.89$ clean AUROC and
falls to chance ($0.52$ to $0.55$) under all evaluated regeneration
attacks. We therefore focus our main comparisons on in-generation
watermarks designed for discrete-token AR models, while retaining DWT-DCT
as a post-hoc reference.

\paragraph{In-generation watermarks for AR image models.}
Most token-level AR watermarks differ in how they partition the codebook.
IndexMark~\cite{tong2025indexmark} computes a maximum-weight perfect
matching on the codebook $k$NN graph to pair each red token with its
closest green neighbor ($\gamma{=}0.5$), combining sampling-time
replacement with an index-reconstruction module against perturbations.
ClusterMark~\cite{lukovnikov2026clustermark} partitions the codebook via
$k$-means and hashes the preceding token's cluster ID so tokens in the
same cluster share colors and transitions. BitMark~\cite{kerner2025bitmark}
watermarks individual bits in bitwise AR generators, and
UniMark~\cite{yilmaz2026unimark} supports next-token and next-scale models
with BCH-coded multi-bit payloads. Yet these methods either omit generation
context altogether or hard-quantize a single preceding token into a
discrete cluster label, leaving a step-function boundary vulnerable to
cross-cluster token drift and cluster-level spoofing.

A complementary direction improves robustness at the tokenizer level
rather than the partition rule: WMAR~\cite{jovanovic2025wmar} retains a
KGW-style split while fine-tuning the VQ autoencoder for reverse cycle
consistency and adding geometric synchronization so edited images
re-encode close to their original tokens. Whereas WMAR stabilizes the
recovered tokens, our approach stabilizes the greenlist rule evaluated on
drifted tokens. Crucially, WMAR requires retraining each VQ tokenizer
separately, whereas our signal network keeps the tokenizer frozen and
transfers across model scales and resolutions without retraining
(\cref{sec:scaling}).

\begin{figure*}[t]
\centering
\includegraphics[width=\textwidth]{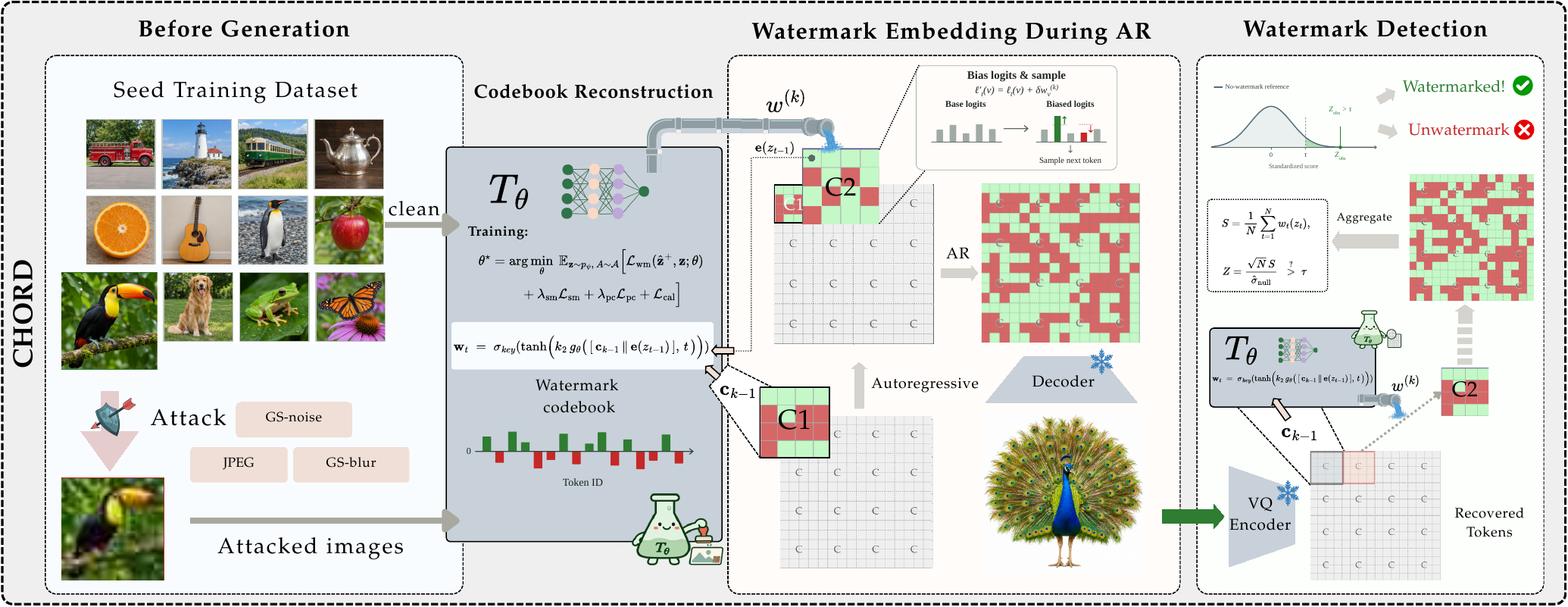}
\caption{\textbf{Overview of \method{}.} During generation, the signal
network $T_\theta$ combines the normalized mean codebook embedding
$\mathbf{c}_{k-1}$ of the preceding token chunk (seeded by the secret key
for the first chunk) with the preceding token embedding
$\mathbf{e}(z_{t-1})$ to predict a vocabulary bias
$\mathbf{w}_t\!\in\![-1,\!+1]^V$, which shifts the frozen generator logits
by $\delta\mathbf{w}_t$. At verification, an image is re-encoded into VQ
tokens, $T_\theta$ is replayed on the recovered sequence, and the average
score $\frac{1}{N}\sum_t w_t(z_t)$ is compared against a calibrated
threshold.}
\label{fig:overview}
\end{figure*}

\paragraph{Watermark attacks.}
Beyond standard image corruptions and regeneration, rule-extraction
attacks directly target the greenlist partition~\cite{zhang2024watermark-stealing}.
First, passive frequency counting tallies token occurrences across
watermarked images; this is far more practical for VQ codebooks than for
text vocabularies because the codebook is much smaller and
nearest-neighbor token replacement in VQ space preserves visual appearance
without introducing linguistic
artifacts~\cite{krishna2023paraphrasing,sadasivan2023cantdetect,piet2023mark}.
Second, a context-aware attacker can group tokens by their conditioning
context to estimate context-specific greenlists. We evaluate both attacks
in \cref{sec:removal}, and extend context-aware inference to continuous
representations by clustering our chunk summaries before estimating
per-cluster partitions.

\section{Method}
\label{sec:method}

\subsection{Preliminaries and threat model}

A token-based AR image generator samples visual tokens
$\mathbf{z}{=}(z_1,\ldots,z_N)$ from a codebook of size $V$
($V{=}16\,384$ for LlamaGen). At step $t$, a \emph{soft} watermark shifts
pre-softmax logits $\ell_t(\cdot\,|\,z_{<t})\in\mathbb{R}^V$ via a bias
$\mathbf{w}_t\in[-1,+1]^V$ scaled by $\delta$:
\begin{equation}
\tilde p_t(z\mid z_{<t}) \;\propto\;
    \exp\!\big(\ell_t(z) + \delta\, w_t(z)\big).
\label{eq:softwm}
\end{equation}
Positive entries of $\mathbf{w}_t$ define the step-$t$ greenlist; at
verification, the detector scores an image by
$S(\mathbf{z})=\tfrac{1}{N}\sum_t w_t(z_t)$ against a calibrated
threshold. A scheme is thus specified by how it constructs $\mathbf{w}_t$.

\emph{Threat model.} Primarily, a benign \emph{editor} applies a
semantic-preserving transformation $A$ to a watermarked image $x$, and
the detector must still verify $A(x)$. Secondarily (\cref{sec:removal}),
a black-box \emph{attacker} queries the generator and knows the watermark
family and VQ codebook, without access to the secret key, signal-network
weights $\theta$ of $T_\theta$, or model gradients, aiming to infer the
greenlist rule or remove the watermark while preserving image quality.

\subsection{Overview of \method{}: design and architecture}
\label{sec:design}

At verification, the detector observes re-encoded tokens
$\hat{\mathbf{z}}=\mathrm{VQEnc}(A(x))$, which differ from the generated
sequence $\mathbf{z}$ under token drift. Detection succeeds only if the
bias replayed on $\hat{\mathbf{z}}$ remains aligned with the bias applied
to $\mathbf{z}$. Denoting the conditioning variable at step $t$ by $u_t$,
this imposes two competing requirements:
\begin{itemize}\itemsep1pt\parskip0pt\topsep2pt
\item[] \textbf{(R1) Edit stability.} $u_t(\hat{\mathbf{z}}) \approx
u_t(\mathbf{z})$ whenever $A$ preserves visual content, despite widespread
token changes.
\item[] \textbf{(R2) Trajectory dependence.} $u_t$ must vary across
images and positions so that no fixed partition can be estimated from
aggregate token frequencies.
\end{itemize}
No discrete $u_t$ can satisfy both requirements. Exact prefix hashing
satisfies (R2) by construction but violates (R1), because any altered
context token completely randomizes the hash output. Conversely, a fixed
global split satisfies (R1) trivially but fails (R2). Resolving this
conflict requires a continuous condition $u_t$ paired with a smooth
mapping to $\mathbf{w}_t$ (\cref{fig:overview}).

\paragraph{Chunk representation.} We instantiate $u_t$ via a pooled
codebook summary of the preceding chunk of $C{=}32$ tokens. Let
$\mathbf{e}(z)\in\mathbb{R}^{d}$ denote the $d$-dimensional
unit-normalized codebook embedding of token $z$ ($d{=}8$ for LlamaGen),
and let $k{=}\lfloor t/C\rfloor$ index the current chunk. The preceding
chunk summary is defined as:
\begin{equation}
\mathbf{c}_{k-1} \;=\;
  \frac{\sum_{j\in\mathcal{I}_{k-1}}\mathbf{e}(z_j)}
       {\big\|\sum_{j\in\mathcal{I}_{k-1}}\mathbf{e}(z_j)\big\|_2},
\qquad \mathcal{I}_{k-1}=[\,(k{-}1)C,\,kC\,),
\label{eq:chunkrep}
\end{equation}
where $\mathbf{c}_{0}{=}\textsc{StartEmbed}(\textit{key})$ seeds the first
chunk using the secret key. This pooled summary satisfies both design
desiderata. First, content-preserving edits typically replace tokens with
nearby entries in VQ embedding space, so averaging over $C$ tokens cancels
out local quantization shifts and keeps $\mathbf{c}_{k-1}$ stable (R1).
Second, distinct images depict different visual regions and draw tokens
from different parts of the codebook, ensuring that $\mathbf{c}_{k-1}$
varies across images and chunks (R2).

\paragraph{Signal network.} A lightweight signal network $T_\theta$ maps
the conditioning context to vocabulary biases. Implemented as an
attention-free residual MLP $g_\theta$ (a two-layer input projection,
three width-$512$ residual blocks, and a size-$V$ linear head producing
pre-activation scores), $T_\theta$ concatenates $\mathbf{c}_{k-1}$ with
a local token embedding $\mathbf{e}(z_{t-1})$ at step $t$, adds a learned
positional embedding, applies $\tanh$ with sharpness $k_2{=}2$, and
permutes the output via a key-seeded $\sigma_{\textit{key}}$:
\begin{equation}
  \mathbf{w}_t = \sigma_{\textit{key}}\!\Big(\!\tanh\!\big(k_2\,
  g_\theta([\,\mathbf{c}_{k-1}\,\|\,\mathbf{e}(z_{t-1})\,],t)\big)\!\Big)
  \!\in\![-1,\!+1]^V\!,
\label{eq:signal}
\end{equation}
where $\sigma_{\textit{key}}$ is the identity for the base key and an
independent pseudorandom permutation per user key (Appendix~A.1, C.7),
keying every step $t\in\{1,\dots,N\}$ without retraining. While $\mathbf{e}(z_{t-1})$ and
$t$ update $\mathbf{w}_t$ per step to aid decorrelation, the shared chunk
summary $\mathbf{c}_{k-1}$ preserves the signal when edits corrupt
$z_{t-1}$. Of $T_\theta$'s $9.44$M parameters ($8.5\%$ of LlamaGen-B,
$2.8\%$ of L), $8.40$M lie in the $V{\times}512$ head, decoupling
$T_\theta$ from generator scale (\cref{sec:scaling}).

\subsection{Training}
\label{sec:losses}

We train $T_\theta$ on clean ImageNet token sequences from the frozen
generator, pairing each unwatermarked reference with a watermarked
sequence sampled under~\cref{eq:softwm} via differentiable soft
re-tokenization and augmenting both with random JPEG, noise, or blur.
Matching~\cref{sec:design}, the training objective is:
\begin{equation}
\label{eq:loss}
\mathcal{L} =
\underbrace{\mathcal{L}_\mathrm{wm}}_{\text{detect.\ / fidelity}}
+ \lambda_\mathrm{sm}\underbrace{\mathcal{L}_\mathrm{sm}}_{\text{(R1)}}
+ \lambda_\mathrm{pc}\underbrace{\lpc}_{\text{(R2)}}
+ \underbrace{\mathcal{L}_\mathrm{cal}}_{\text{scale}} ,
\end{equation}
with $\lambda_\mathrm{sm}{=}0.1$ and $\lambda_\mathrm{pc}{=}2.5$. Here,
$\mathcal{L}_\mathrm{wm}$ pairs a position-level binary cross-entropy
detection loss with an LPIPS${+}$MSE perceptual fidelity loss, while
$\mathcal{L}_\mathrm{cal}$ centers $\mathbf{w}_t$, prevents $\tanh$
saturation, and maintains a target green fraction $\gamma{=}0.25$
(Appendix~D.2), regulating only signal scale rather than partition
structure (full definitions and schedules are in Appendix~A.2).

\emph{Smoothness (R1).} $\mathcal{L}_\mathrm{sm}=\mathcal{L}_s+
\mathcal{L}_\mathrm{cb}$ enforces smoothness in both input and codeword
spaces. Over pairs $(i,j)$ of subsampled batch positions with context
vectors $\mathbf{u}_i=[\,\mathbf{c}_{k(i)-1}\,\|\,\mathbf{e}(z_{t(i)-1})\,]$
and pre-$\tanh$ outputs $\tilde{\mathbf{w}}_i=g_\theta(\mathbf{u}_i,t(i))$,
\begin{equation}
\mathcal{L}_s = \mathbb{E}_{i,j}\Big[\;\big\lvert
 \cos\!\big(\tilde{\mathbf{w}}_i,\tilde{\mathbf{w}}_j\big)
 - \tanh\!\big(k_1(\cos(\mathbf{u}_i,\mathbf{u}_j) - \overline{\cos})\big)
\big\rvert\;\Big] ,
\label{eq:lsim}
\end{equation}
where $\overline{\cos}=\mathbb{E}_{i',j'}[\cos(\mathbf{u}_{i'},\mathbf{u}_{j'})]$
is the batch mean similarity and $k_1{=}20$ scales temperature, mapping
similar contexts to similar greenlists while separating distinct ones.
Over codewords, $\mathcal{L}_\mathrm{cb}$ smooths $\mathbf{w}_t$ across
nearest codebook neighbors, a continuous counterpart to ClusterMark's hard
$k$-means~\cite{lukovnikov2026clustermark}.

\emph{Decorrelation (R2).} To prevent collapse onto a static greenlist,
for each sequence we uniformly sample $n{=}16$ positions
$\{t_1,\dots,t_n\}\subset\{1,\dots,N\}$ and penalize the mean bias that
$t_a$ assigns to $t_b$'s top-$K$ favored tokens:
\begin{equation}
\lpc = \frac{1}{n(n{-}1)}\sum_{a\neq b}
 \ \mathbb{E}\!\left[\frac{1}{K}\textstyle\sum_{z\in\mathrm{top}K(\mathbf{w}_{t_b})}
   w_{t_a}(z)\right] ,
\label{eq:lpc}
\end{equation}
averaged over batch sequences and random $n$-subsets. If all positions
shared one greenlist, their top-$K$ sets would coincide and incur a large
penalty; as shown in~\cref{sec:ablation}, \lpc{} directly governs the
robustness and secrecy trade-off, as removing it drives the learned rule
toward a static global split.

Loss weights are fixed once on LlamaGen-B and reused unchanged on
LlamaGen-L and across all ablations. Unless stated otherwise, all
experiments use $\delta{=}4$, $\gamma{=}0.25$, and $C{=}32$.

\section{Experiments}
\label{sec:experiments}

\subsection{Setup}
\label{sec:setup}

\paragraph{Generators.}
Public class-conditional LlamaGen checkpoints at two sizes ($111$M
c2i\_B, $343$M c2i\_L) on $256{\times}256$ images, with a
$16{\times}16$ VQ grid ($N{=}256$) and $V{=}16\,384$ codebook.
Cross-resolution transfer to $384{\times}384$ checkpoints and
cross-tokenizer evaluation on RAR-XL~\cite{yu2025rar} ($V{=}1\,024$,
$d{=}256$) are reported in Appendix~D.5 and D.6.

\paragraph{Baselines.}
\textbf{KGW-$h$} seeds the greenlist by hashing the $h$ preceding tokens
with the key, for $h\!\in\!\{0,1,2,4\}$; $h{=}0$ is the global split of
the unigram variant~\cite{zhao2024provable} and $h{=}1$ the original
rule~\cite{kirchenbauer2023kgw}. \textbf{Chunk-hash} keeps our chunk
structure but derives each chunk's partition by \emph{exactly} hashing
the preceding chunk's discrete token tuple instead of pooling its
embeddings (a training-free continuous counterpart that quantizes and
hashes $\mathbf{c}_{k-1}$ directly on LlamaGen-B is also evaluated in
Appendix~D.6). We further compare against
ClusterMark~\cite{lukovnikov2026clustermark},
IndexMark~\cite{tong2025indexmark}, and a pixel-domain DWT-DCT baseline.
All token-level schemes share the detector template
$S(\mathbf{z})=\tfrac{1}{N}\sum_t w_t(z_t)$; the hash baselines run at
$\delta{=}8$ against our $\delta{=}4$ to equalize the green-to-red logit
gap. Hyperparameters are in Appendix~A.2.

\paragraph{Edits and protocol.}
We test eleven semantic-preserving edits (DINOv2~\cite{oquab2024dinov2}
cosine ${\geq}0.870$, plus CLIP~\cite{radford2021clip} and
LPIPS~\cite{zhang2018lpips}; Appendix~B.1), grouping rotation and
cropping as \emph{misaligning} edits. We report AUROC and $\TPR@1\%\FPR$
evaluated against unwatermarked images under the same edit ($300$ to $400$
images per condition; default $\delta{=}4, \gamma{=}0.25, C{=}32$, key
$2024$), alongside two rule-extraction attacks (\cref{sec:removal}),
broader image/regeneration benchmarks, and FID/IS
(Appendix~D.7)~\cite{heusel2017fid}.

\subsection{Main results}
\label{sec:main}

\begin{table*}[t]
\centering
\caption{\textbf{Clean detection, edit robustness, and rule secrecy on LlamaGen-B.}
Under \textbf{Clean} and \textbf{Robustness}, cells report
AUROC\,/\,$\TPR@1\%\FPR$ on $400$ test images evaluated against
identically edited negative images (confidence intervals in Appendix~B.2).
Hash baselines use $\delta{=}8$ to match the logit
gap of \method{} at $\delta{=}4$; chunk-hash applies exact hashing to the
same chunks as \method{}. Under \textbf{Secrecy} ($500$-image budget;
\cref{sec:removal}), cells are shaded \colorbox{green!16}{green} if
usable and \colorbox{red!12}{red} if compromised (parentheses show change
from clean AUROC). Bold marks the best AUROC and lowest rule-inference
accuracy.}
\label{tab:edit_detection}
\setlength{\tabcolsep}{4.5pt}
\footnotesize
\begin{tabular}{ll|cccccc}
\toprule
Category & Evaluation & \oc{\method{}} & KGW $h{=}0$ & KGW $h{=}1$ & KGW $h{=}2$ & KGW $h{=}4$ & chunk-hash \\
\midrule
\textbf{Clean} & no edit & \oc{\best{1.000}\,/\,1.000} & \best{1.000}\,/\,1.000 & \best{1.000}\,/\,1.000 & \best{1.000}\,/\,1.000 & \best{1.000}\,/\,0.993 & 0.977\,/\,0.752 \\
\midrule
\multirow{16}{*}{\textbf{Robustness}}
 & \multicolumn{7}{@{}l}{\itshape Aligned edits: compression} \\
 & \quad JPEG $q25$            & \oc{0.988\,/\,0.965} & \best{0.993}\,/\,0.970 & 0.976\,/\,0.838 & 0.902\,/\,0.547 & 0.744\,/\,0.242 & 0.797\,/\,0.215 \\
 & \quad WebP $q25$            & \oc{0.981\,/\,0.900} & \best{0.995}\,/\,0.965 & 0.960\,/\,0.780 & 0.894\,/\,0.580 & 0.719\,/\,0.217 & 0.793\,/\,0.220 \\
 & \multicolumn{7}{@{}l}{\itshape Aligned edits: appearance} \\
 & \quad color jitter          & \oc{0.989\,/\,0.963} & \best{0.994}\,/\,0.963 & 0.979\,/\,0.885 & 0.913\,/\,0.675 & 0.783\,/\,0.400 & 0.794\,/\,0.233 \\
 & \quad brightness            & \oc{0.998\,/\,0.995} & 0.999\,/\,0.995 & \best{1.000}\,/\,0.998 & 0.997\,/\,0.955 & 0.969\,/\,0.865 & 0.926\,/\,0.527 \\
 & \quad saturation            & \oc{\best{1.000}\,/\,1.000} & \best{1.000}\,/\,1.000 & \best{1.000}\,/\,1.000 & \best{1.000}\,/\,0.995 & 0.997\,/\,0.973 & 0.977\,/\,0.713 \\
 & \multicolumn{7}{@{}l}{\itshape Aligned edits: signal-level} \\
 & \quad noise $\sigma{=}0.05$ & \oc{0.993\,/\,0.902} & \best{0.997}\,/\,0.978 & 0.985\,/\,0.885 & 0.949\,/\,0.797 & 0.827\,/\,0.448 & 0.760\,/\,0.175 \\
 & \quad blur $r{=}2$          & \oc{\best{1.000}\,/\,0.995} & 0.998\,/\,0.998 & 0.997\,/\,0.980 & 0.987\,/\,0.873 & 0.916\,/\,0.615 & 0.883\,/\,0.412 \\
 & \quad resize $50\%$         & \oc{\best{1.000}\,/\,1.000} & \best{1.000}\,/\,0.998 & 0.998\,/\,0.988 & 0.984\,/\,0.895 & 0.936\,/\,0.693 & 0.897\,/\,0.403 \\
\cmidrule(lr){2-8}
 & \quad \textit{mean (aligned)} & \oc{0.993\,/\,0.965} & \best{0.997}\,/\,0.983 & 0.987\,/\,0.919 & 0.953\,/\,0.790 & 0.861\,/\,0.557 & 0.853\,/\,0.362 \\
\cmidrule(lr){2-8}
 & \multicolumn{7}{@{}l}{\itshape Misaligning edits} \\
 & \quad rotate $5^\circ$      & \oc{\best{0.822}\,/\,0.207} & 0.811\,/\,0.160 & 0.630\,/\,0.022 & 0.536\,/\,0.007 & 0.484\,/\,0.010 & 0.512\,/\,0.005 \\
 & \quad crop $5\%$            & \oc{\best{0.961}\,/\,0.682} & 0.957\,/\,0.647 & 0.804\,/\,0.323 & 0.716\,/\,0.160 & 0.595\,/\,0.045 & 0.570\,/\,0.045 \\
\cmidrule(lr){2-8}
 & \quad \textit{mean (misaligning)} & \oc{\best{0.892}\,/\,0.445} & 0.884\,/\,0.404 & 0.717\,/\,0.173 & 0.626\,/\,0.084 & 0.540\,/\,0.028 & 0.541\,/\,0.025 \\
\midrule
\multirow{3}{*}{\textbf{Secrecy}}
 & frequency-removal AUROC $\uparrow$ & \rsurv{\best{0.943}\,\loss{0.057}} & \rbrk{0.015\,\loss{0.985}} & --- & --- & --- & --- \\
 & rule-inference acc.\ $\downarrow$ & \rsurv{\best{0.536}} & \rbrk{0.986} & n/e & n/e & n/e & \rbrk{0.866} \\
 & \quad (coverage) & \oc{1.000} & 1.000 & 0.004 & 0.004 & 0.004 & 0.125 \\
\bottomrule
\end{tabular}
\end{table*}

\subsubsection{Detection after semantic-preserving edits}
\label{sec:editdet}

\method{} achieves $0.993$ mean AUROC and $0.965$ mean $\TPR@1\%\FPR$ on
aligned edits, closely matching global KGW ($h{=}0$;
\cref{tab:edit_detection}). Under misaligning edits, where token drift is
largest, \method{} matches the static global ceiling ($h{=}0$) within
sampling variation while outperforming one-token KGW ($h{=}1$) by $+0.193$
AUROC on $5^\circ$ rotation and $+0.158$ on $5\%$ cropping, with paired
bootstrap intervals placing both gains well above zero (Appendix~B.2).
Across both edit groups, KGW degrades steadily as $h$ grows, falling near
chance on misaligning edits at $h{=}4$ ($0.540$).

\begin{figure*}[t]
\centering
\includegraphics[width=\textwidth]{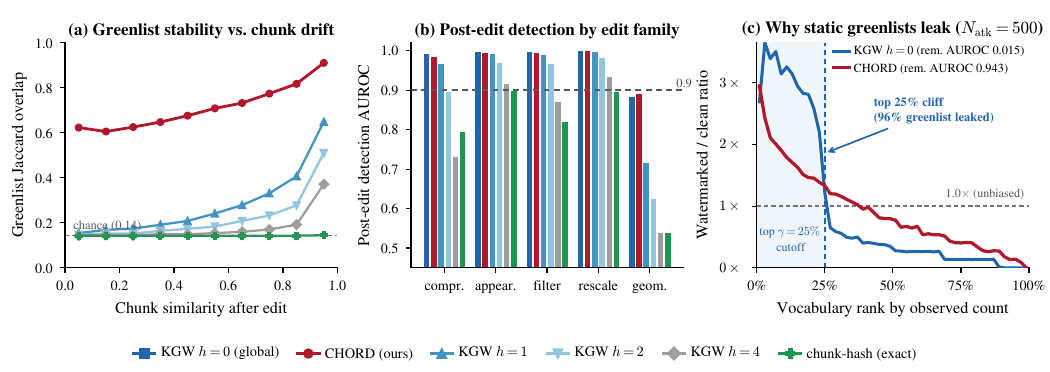}
\caption{\textbf{Greenlist stability, post-edit detection, and static greenlist leakage.}
\textbf{(a)}~Pre- and post-edit greenlist Jaccard overlap binned by chunk
similarity ($76\,800$ position pairs; $h{=}0$ is constant at $1.0$).
\textbf{(b)}~Post-edit detection AUROC across edit families.
\textbf{(c)}~Watermarked-to-clean token frequency ratio sorted by observed
token count over $N_\mathrm{atk}{=}500$ images (\cref{sec:freqremoval}):
global KGW ($h{=}0$) exhibits a sharp ${\sim}3\times$ cliff at the top
$\gamma{=}25\%$ cutoff that leaks $96\%$ of its greenlist (removal AUROC
$0.015$), whereas \method{} decays smoothly without a separable cutoff
(retaining $0.943$; raw counts in Appendix~C.2).}
\label{fig:edit_main}
\end{figure*}

\method{} retains an AUROC of $0.943$ under frequency-based removal.
At full coverage, the clustering-based attack infers \method{}'s rules
with near-chance accuracy ($0.536$). For global KGW, the removal attack
drives AUROC to $0.015$ once its fixed greenlist is recovered, leaving
the original detector unreliable in this threat model.
\cref{fig:edit_main}c shows the token statistics this separation
rests on.

\subsubsection{Stability of the conditioning signal}
\label{sec:consistency}

\begin{table}[t]
\centering
\caption{\textbf{Token drift and greenlist stability}
(LlamaGen-B, $400$ images, $16$ positions each). Drift measures the
fraction of changed token IDs. Cosine similarity compares pre- and
post-edit chunk representations (\cref{eq:chunkrep}). The remaining
columns report greenlist Jaccard similarity, with a chance reference
of $0.143$ at $\gamma{=}0.25$. The cross-source row pairs chunks from
different images.}
\label{tab:consistency}
\setlength{\tabcolsep}{3.2pt}
\scriptsize
\begin{tabular}{l|cc|cccc c}
\toprule
 & & & \multicolumn{5}{c}{greenlist Jaccard $\uparrow$} \\
\cmidrule(lr){4-8}
Edit & drift & cos & ours & $h{=}1$ & $h{=}2$ & $h{=}4$ & c-hash \\
\midrule
saturation          & 0.100 & 0.994 & \best{0.973} & 0.914 & 0.843 & 0.745 & 0.333 \\
brightness          & 0.251 & 0.960 & \best{0.933} & 0.786 & 0.689 & 0.562 & 0.247 \\
resize $50\%$       & 0.357 & 0.961 & \best{0.922} & 0.692 & 0.542 & 0.385 & 0.157 \\
JPEG $q50$          & 0.407 & 0.959 & \best{0.911} & 0.661 & 0.510 & 0.348 & 0.145 \\
blur $r{=}2$        & 0.410 & 0.929 & \best{0.896} & 0.656 & 0.488 & 0.326 & 0.146 \\
colour jitter       & 0.539 & 0.890 & \best{0.882} & 0.545 & 0.400 & 0.267 & 0.146 \\
noise $\sigma{=}.05$& 0.544 & 0.881 & \best{0.838} & 0.545 & 0.407 & 0.272 & 0.144 \\
JPEG $q25$          & 0.607 & 0.902 & \best{0.867} & 0.490 & 0.344 & 0.222 & 0.143 \\
WebP $q25$          & 0.632 & 0.904 & \best{0.858} & 0.461 & 0.313 & 0.204 & 0.143 \\
crop $5\%$          & 0.854 & 0.940 & \best{0.826} & 0.271 & 0.203 & 0.166 & 0.143 \\
rotate $5^\circ$    & 0.957 & 0.899 & \best{0.776} & 0.234 & 0.183 & 0.146 & 0.143 \\
\midrule
\textit{cross-source} & 1.000 & 0.160 & 0.637 & 0.143 & 0.143 & 0.143 & 0.143 \\
\bottomrule
\end{tabular}
\end{table}

Chunk representations remain similar after editing despite substantial
token drift (\cref{tab:consistency}). \method{} retains more greenlist
overlap than the exact-hash baselines, including the control with the
same chunk structure. Its overlap rises with chunk similarity, while
the exact-hash baselines recover overlap primarily in the
highest-similarity bins (\cref{fig:edit_main}a).

\method{}'s greenlists have a mean Jaccard similarity of $0.637$ when
paired across different images, reflecting the smoothness enforced by
$\mathcal{L}_\mathrm{sm}$: crucially, this pairwise overlap does not
concentrate on a fixed subset of tokens across trajectories, as aggregate
token frequencies remain unbiased and static top-$\lceil\gamma V\rceil$
extraction fails to remove or forge the watermark
(\cref{fig:edit_main}c, \cref{sec:removal}, Appendix~C.5). To account
for baseline overlap when comparing greenlist recovery, we normalize the
Jaccard similarity $J$ between pre- and post-edit greenlists as
$(J-J_\mathrm{floor})/(1-J_\mathrm{floor})$, using each method's mean
cross-source Jaccard as $J_\mathrm{floor}$. A score of $0$ matches the
cross-source overlap, and $1$ indicates identical greenlists. \method{}
leads the exact-hash baselines on all eleven edits under this measure.
Under rotation, the adjusted overlap is $0.38$ for \method{} and $0.11$
for the one-token KGW baseline (Appendix~D.6 further evaluates a
training-free quantized-chunk control alongside cross-tokenizer transfer).

\subsubsection{Clean detection and fidelity}
\label{sec:fid}

On unedited LlamaGen-B outputs (\cref{tab:edit_detection}), \method{}
achieves $1.000$ AUROC and $1.000$ TPR at $1\%$ FPR, matching the
strongest KGW baselines while surpassing exact chunk hashing. Matched-seed
LlamaGen-L generations alongside ClusterMark and IndexMark appear in
\cref{fig:qualitative}, with FID and IS reported in Appendix~D.7.
\begin{figure}[t]
\centering
\setlength{\tabcolsep}{1pt}
\renewcommand{\arraystretch}{0.8}
\scriptsize
\begin{tabular}{@{}cccc@{}}
clean & \method{} & ClusterMark & IndexMark \\[-1pt]
\includegraphics[width=.235\columnwidth]{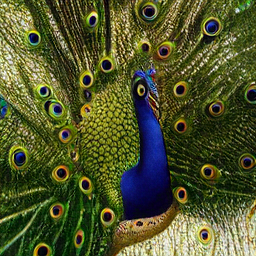} &
\includegraphics[width=.235\columnwidth]{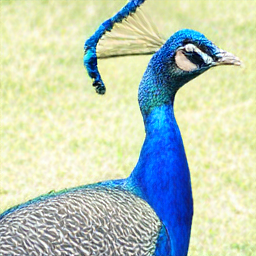} &
\includegraphics[width=.235\columnwidth]{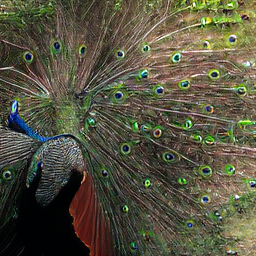} &
\includegraphics[width=.235\columnwidth]{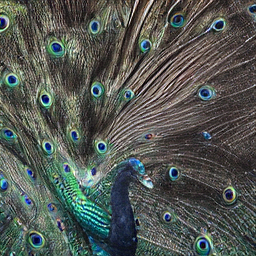} \\[-1pt]
\multicolumn{4}{c}{\scriptsize peacock} \\[1pt]
\includegraphics[width=.235\columnwidth]{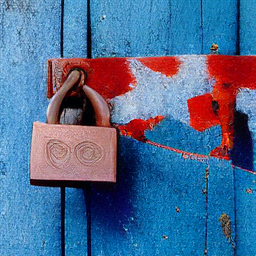} &
\includegraphics[width=.235\columnwidth]{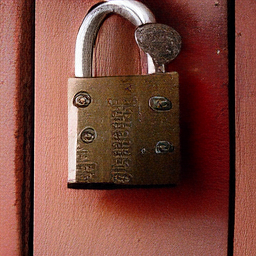} &
\includegraphics[width=.235\columnwidth]{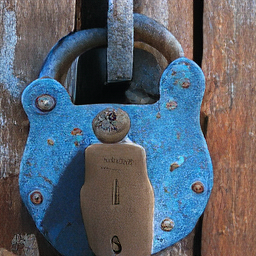} &
\includegraphics[width=.235\columnwidth]{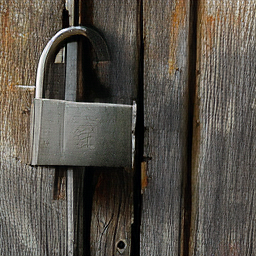} \\[-1pt]
\multicolumn{4}{c}{\scriptsize padlock} \\[1pt]
\includegraphics[width=.235\columnwidth]{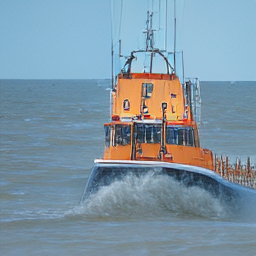} &
\includegraphics[width=.235\columnwidth]{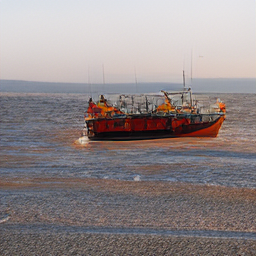} &
\includegraphics[width=.235\columnwidth]{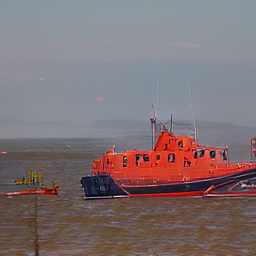} &
\includegraphics[width=.235\columnwidth]{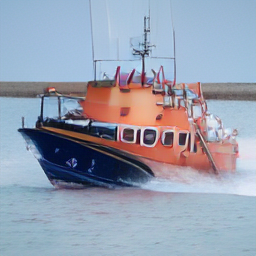} \\[-1pt]
\multicolumn{4}{c}{\scriptsize lifeboat} \\
\end{tabular}
\caption{\textbf{Qualitative comparison on LlamaGen-L.}
Within each row, all methods use the same ImageNet class and random
seed. \method{} uses $\delta{=}4$. Extended comparisons over six
classes and five attacks per class appear in Appendix~B.6.}
\label{fig:qualitative}
\end{figure}

\subsubsection{The broader attack suite}
\label{sec:robust}

This benchmark covers geometric, combined, and regeneration attacks.
\cref{tab:geom} reports LlamaGen-L; the LlamaGen-B counterpart over the
same attacks is in Appendix~B.3, and full strength sweeps at both scales
are in Appendix~B.5.

\begin{table}[t]
\centering
\caption{\textbf{Robustness to image-level and regeneration attacks on
LlamaGen-L} (AUROC; $\delta{=}4$ for \method{}, $\tau{=}0$ for IM).
Bold$+$underlined marks the best token-level AUROC; stars and parentheses
show where \method{} matches/exceeds the best token-level baseline and its
signed margin. Generator-agnostic DWT-DCT~\cite{khan2013digitalwatermarkingimageauthenticationbased}
(on LlamaGen-B) is shown for reference. LlamaGen-B and full 21-attack
evaluations are in Appendix~B.3 and~B.4.}
\label{tab:geom}
\scriptsize
\setlength{\tabcolsep}{2.5pt}
\setlength{\cnumwd}{2.3em}\setlength{\cstarwd}{0.7em}\setlength{\cannwd}{2.6em}
\begin{tabular}{l|cccc|c}
\toprule
 & \multicolumn{4}{c|}{\textbf{LlamaGen-L} (343\,M)} & \textbf{Pixel} \\
\cmidrule(lr){2-5} \cmidrule(lr){6-6}
Attack & \method{} & KGW $h{=}0$ & CM & IM & DWT-DCT \\
\midrule
clean                        & \best{1.000} & \best{1.000} & \best{1.000} & \best{1.000} & 0.893 \\
\midrule
\multicolumn{6}{@{}l}{\itshape Geometric transformations} \\
\quad rot $5^\circ$          & \cwin{0.807}{\gain{0.002}} & 0.805 & 0.791 & 0.622 & 0.741 \\
\quad rot $30^\circ$         & \cwin{0.612}{\gain{0.004}} & 0.606 & 0.608 & 0.582 & 0.562 \\
\quad rot $90^\circ$         & \cwin{0.582}{\gain{0.002}} & 0.579 & 0.580 & 0.518 & 0.331 \\
\quad hflip                  & \cwin{0.763}{\gain{0.038}} & 0.725 & 0.646 & 0.573 & 0.460 \\
\quad vflip                  & \cwin{0.665}{\gain{0.006}} & 0.659 & 0.643 & 0.566 & 0.888 \\
\quad resize $25\%$          & \clos{0.954}{\loss{0.007}} & 0.955 & \textbf{\underline{0.961}} & 0.792 & 0.610 \\
\midrule
\multicolumn{6}{@{}l}{\itshape Compression, noise \& blur} \\
\quad JPEG $q50$             & \cwin{1.000}{\gain{0.000}} & 0.998 & 1.000 & 0.987 & 0.532 \\
\quad noise $\sigma{=}.05$   & \cwin{0.997}{\gain{0.004}} & 0.993 & 0.991 & 0.953 & 0.862 \\
\quad blur $r{=}2$           & \cwin{1.000}{\gain{0.000}} & 1.000 & 1.000 & --- & 0.752 \\
\midrule
\multicolumn{6}{@{}l}{\itshape Combined pipelines} \\
\quad hflip$+$JPEG $q25$     & \cwin{0.596}{\gain{0.007}} & 0.589 & 0.558 & 0.500 & 0.572 \\
\quad rot $5^\circ$$+$crop $5\%$ & \clos{0.815}{\loss{0.059}} & \textbf{\underline{0.874}} & 0.808 & 0.648 & 0.731 \\
\quad resize $50\%$$+$noise  & \cwin{0.993}{\gain{0.003}} & 0.989 & 0.990 & --- & 0.860 \\
\midrule
\multicolumn{6}{@{}l}{\itshape Regeneration} \\
\quad VQ regen ($k{=}5$)     & \clos{0.740}{\loss{0.211}} & 0.635 & \textbf{\underline{0.951}} & 0.472 & 0.525 \\
\quad Diff.\ regen ($s{=}.5$) & \cwin{0.603}{\gain{0.006}} & 0.584 & 0.597 & 0.508 & 0.550 \\
\bottomrule
\end{tabular}
\end{table}

\method{} leads the token-level baselines in most geometric and
combined settings at both scales, with its largest gain under
horizontal flipping. All evaluated token-level methods retain high
AUROC under moderate compression, noise, and blur.

Under SD img2img regeneration, \method{} substantially leads on
LlamaGen-B ($+0.105$ AUROC; Appendix~B.3) and matches or edges out
the strongest baseline across strength sweeps on LlamaGen-L
(Appendix~B.5). ClusterMark leads under VQ-$k$NN remapping,
consistent with its cluster-level coloring, while DWT-DCT falls near
chance on both regeneration tests (Appendix~B.4).
\subsection{Secrecy of the conditioning rule}
\label{sec:removal}

Because a trivial static greenlist maximizes edit stability, we evaluate
rule secrecy as the complementary axis: how accurately an adversary can
infer or strip the partition from watermarked outputs.

\subsubsection{Frequency-counting removal}
\label{sec:freqremoval}

\paragraph{Attack setup.}
The attacker queries $N_\mathrm{atk}{=}500$ watermarked images and tallies
empirical token counts across the $V{=}16\,384$ codebook. Under a fixed
global split, green tokens are consistently inflated, so taking the
top-$\lceil\gamma V\rceil$ most frequent tokens recovers the greenlist
($0.96$ Jaccard at $\delta{=}4$; Appendix~C.1). Replacing each
inferred green token with its nearest non-green codebook neighbor removes
the watermark with minimal perceptual distortion~\cite{razavi2019vqvae2}.

\begin{figure}[t]
\centering
\includegraphics[width=\columnwidth]{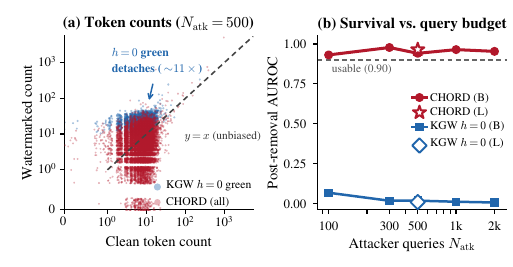}
\caption{\textbf{Frequency-counting watermark removal.}
\textbf{(a)}~Per-token watermarked vs.\ clean counts over
$N_\mathrm{atk}{=}500$ LlamaGen-B images ($\delta{=}4$): global KGW
($h{=}0$) inflates its fixed greenlist above $y{=}x$, whereas \method{}
stays unbiased along the diagonal.
\textbf{(b)}~Post-removal detection AUROC across query budgets
$N_\mathrm{atk}\!\in\![100,2000]$: global KGW ($h{=}0$) collapses near
zero on both LlamaGen-B and L, while \method{} retains $0.93$ to $0.98$
on B and $0.965$ on L (full baseline breakdown in Appendix~C.1).}
\label{fig:removal_main}
\end{figure}

\paragraph{Results.}
Global KGW ($h{=}0$) collapses below chance because $k$NN replacement
systematically steers tokens into its true redlist
(\cref{fig:removal_main}b), already failing at $N_\mathrm{atk}{=}100$
(additional single-split and clustered baselines are in Appendix~C.1).
By contrast, \method{} retains $0.93$ to $0.98$
AUROC across $N_\mathrm{atk}\in[100,2000]$ on LlamaGen-B and $0.965$ on
LlamaGen-L because trajectory-dependent partitions keep aggregate token
counts unbiased along $y{=}x$ (\cref{fig:removal_main}a).

\subsubsection{Rule inference under a context-aware attacker}
\label{sec:ruleinfer}

To test context-dependent schemes ($h{\geq}1$ and \method{}), a stronger
attacker groups tokens by their conditioning context and estimates a
per-group greenlist: exact-hash schemes are bucketed by exact hash seed,
whereas \method{}'s continuous chunk summaries are quantized via spherical
$k$-means ($k{=}64$; granularity sweeps in Appendix~C.3).

\cref{tab:edit_detection} (bottom) reports balanced accuracy alongside
context coverage. For $h{=}0$, a single bucket covers all tokens and leaks
the split ($0.986$). For $h\!\in\!\{1,2,4\}$, $V^h$ contexts ($V{=}16\,384$)
leave every bucket below the sample floor except the key-seeded start
token ($0.004$ coverage), making their rules unestimable rather than
empirically tested at $500$ images. Chunk-hash leaks its key-seeded first
chunk ($0.866$ at $0.125$ coverage). Only \method{} holds the attacker
near chance ($0.536$) at \emph{full} $1.000$ coverage, as each $k$-means
cell averages over too many distinct continuous partitions for a
consistent greenlist to emerge; this near-chance bound persists across a
$20\times$ range of query budgets and $16\times$ range of cluster counts
(Appendix~C.3, with cluster-aware removal, spoofing, and continuous
neural surrogate attacks evaluated in Appendix~C.4 to~C.6).

\subsection{Ablations and scaling}
\label{sec:ablation}
\label{sec:scaling}

We retrain $T_\theta$ across eight design variants on LlamaGen-B
(Appendix~D.1). Removing \lpc{} is the only change that breaks
removal resistance ($0.943{\to}0.285$ AUROC) while raising image-attack
AUROC to $0.949$ as positions collapse toward a near-global split,
demonstrating that $\lambda_\mathrm{pc}$ directly traverses the
robustness and secrecy trade-off within one architecture
(Appendix~D.3). Across the remaining seven variants, post-removal AUROC
stays $\geq 0.916$, while adding the local token branch ($0.796{\to}0.822$
on rotation and $0.677{\to}0.742$ on horizontal flip), adopting a $C{=}32$
chunk width over $C{=}16$ ($+0.028$ mean AUROC), and training with
differentiable attack augmentation ($+0.050$ mean AUROC) each contribute
clear gains in spatial robustness. Retraining on LlamaGen-L with identical
hyperparameters preserves both robustness and secrecy ($0.965$ removed
AUROC), and the $256^2$-trained $T_\theta$ transfers zero-shot to $384^2$
($576$ tokens) via linear positional interpolation while retaining strong
post-edit detection ($0.941$ aligned mean and $0.869$ rotation AUROC on
LlamaGen-L-$384$, plus $0.897$ on B-$384$; Appendix~D.5, with
cross-tokenizer results in Appendix~D.6).

\section{Discussion and Conclusion}
\label{sec:discussion}

Beyond surviving benign edits and frequency counting, \method{} offers
practical properties for real-world deployment. Because a single trained
$T_\theta$ keeps the VQ tokenizer frozen, it transfers zero-shot across
resolutions (\cref{sec:scaling}; Appendix~D.5), supports arbitrary secret
keys via per-step vocabulary permutation without retraining
(Appendix~C.7), and blocks watermark spoofing at a $2.0\%$ pass rate
(Appendix~C.5). Several directions remain open for future work, including
distilling lighter output heads to reduce per-step generation latency
(Appendix~D.8), pairing \method{} with geometric synchronization to
further improve recall under severe spatial transformations
(Appendix~B.5), and extending our zero-bit design to multi-bit payloads
(Appendix~D.6).

We presented \method{}, an in-generation AR image watermark
that resolves the tension between post-edit greenlist recovery and rule
secrecy by smoothly mapping continuous chunk summaries to decorrelated
vocabulary biases. On LlamaGen-B, \method{} matches global KGW on aligned
edits, outperforms one-token KGW on rotation and cropping by $0.175$ mean
AUROC and $0.272$ mean $\TPR@1\%\FPR$, and leads published AR watermarks
under diffusion regeneration, whereas exact chunk hashing falls to
near-chance overlap. At a $500$-image query budget, \method{} retains
$0.943$ AUROC under frequency-based removal and holds context-clustering
rule inference near chance ($0.536$ at full coverage), enabling robust and
secret provenance for autoregressive image generation.
{\small
\bibliographystyle{ieeenat_fullname}
\bibliography{refs}
}

\clearpage
\appendix

\renewcommand{\topfraction}{0.95}
\renewcommand{\bottomfraction}{0.90}
\renewcommand{\textfraction}{0.05}
\renewcommand{\floatpagefraction}{0.85}
\renewcommand{\dbltopfraction}{0.95}
\renewcommand{\dblfloatpagefraction}{0.85}

\section{Algorithm and Implementation Details}
\label{sec:supp_algo_impl}

\subsection{Generation and verification in full}
\label{supp:algorithm}

\cref{alg:chunkvisir} states the two routines that define \method{} in
full: \textsc{Generate}, which biases the frozen generator's logits by
the chunk-conditioned signal at every step, and \textsc{Verify}, which
re-tokenizes a possibly edited image and replays the same signal on the
tokens it recovers. Both call the same \textsc{ChunkSignal}, so the
detector needs no side information beyond the key, as the condition it
scores against is recomputed directly from the recovered visual token
sequence $(\hat z_1,\dots,\hat z_N)$ without requiring the class label
$c$. At the sequence and first-chunk boundaries ($t{=}1$ for the local
branch and $k{=}0$, i.e., $t\in\{1,\dots,C\}$, for the chunk branch),
\textsc{ChunkSignal} substitutes $\textsc{StartEmbed}(\textit{key})$,
ensuring identical boundary handling during generation and verification.

\begin{algorithm}[!t]
\caption{Chunk-conditioned soft watermarking.
\textsc{Generate} produces a watermarked token sequence from a class
condition $c$; \textsc{Verify} returns the mean signal value over the $N$
visual tokens, compared against a threshold calibrated at a target FPR on
negative images. $T_\theta$ is the trained signal network
of~Sec.~3.2, shared by both routines.}
\label{alg:chunkvisir}
\footnotesize
\begin{algorithmic}[1]
\Function{ChunkSignal}{$z_{<t}$, $t$, $\theta$, \textit{key}}
    \State $k \gets \lfloor (t-1)/C\rfloor$ \Comment{$0$-based chunk index}
    \If{$k = 0$} \Comment{first chunk ($t\le C$): seed from key}
        \State $\mathbf{c} \gets \textsc{StartEmbed}(\textit{key})$
    \Else \Comment{pooled previous chunk}
        \State $\mathbf{c} \gets \mathrm{normalize}\big(\mathrm{mean}(\{\mathbf{e}(z_j)\}_{j=(k-1)C+1}^{kC})\big)$
    \EndIf
    \If{$t = 1$} \Comment{first token: seed local branch}
        \State $\mathbf{l} \gets \textsc{StartEmbed}(\textit{key})$
    \Else
        \State $\mathbf{l} \gets \mathbf{e}(z_{t-1})$
    \EndIf
    \State \Return $\mathbf{w}_t \gets \sigma_{\textit{key}}\!\Big(\tanh\!\big(k_2\,g_\theta([\,\mathbf{c}\,\|\,\mathbf{l}\,],\,t)\big)\Big) \in[-1,+1]^V$
\EndFunction
\Statex
\Function{Generate}{class $c$, $\theta$, $\delta$, \textit{key}}
    \For{$t = 1,\ldots,N$}
        \State $\ell_t \gets \text{Generator}(c, z_{<t})$ \Comment{frozen AR logits}
        \State $\mathbf{w}_t \gets \textsc{ChunkSignal}(z_{<t},t,\theta,\textit{key})$
        \State $\tilde p_t(z) \propto \exp\!\big(\ell_t(z) + \delta\, w_t(z)\big)$
        \State $z_t \sim \tilde p_t$
    \EndFor
    \State \Return $(z_1,\ldots,z_N)$
\EndFunction
\Statex
\Function{Verify}{image $x$, $\theta$, \textit{key}}
    \State $(\hat z_1,\ldots,\hat z_N) \gets \text{VQEncoder}(x)$ \Comment{re-tokenize $x$}
    \For{$t = 1,\ldots,N$}
        \State $\mathbf{w}_t \gets \textsc{ChunkSignal}(\hat z_{<t},t,\theta,\textit{key})$ \Comment{replay on $\hat z_{<t}$}
    \EndFor
    \State \Return $S \gets \tfrac{1}{N}\sum_{t=1}^{N} w_t(\hat z_t)$
\EndFunction
\end{algorithmic}
\end{algorithm}

\subsection{Implementation details}
\label{supp:implementation}

\emph{Calibration terms.} $\mathcal{L}_\mathrm{cal}$
of~Eq.~(5) is $\lambda_n\mathcal{L}_n + \lambda_\mathrm{gf}\lgf
+ \lambda_\mathrm{ent}\lent$. Writing $\tilde{\mathbf{w}}$ for the
pre-$\tanh$ output, $\mathcal{L}_n = |\overline{\tilde{w}}| +
\mathbb{E}\big|R - |\tilde{w}|\big|$ with $R{=}1$ centers the signal
over the vocabulary and holds its magnitude away from zero; $\lent =
-\mathbb{E}\log(1 - w^2)$ keeps the post-$\tanh$ output away from
saturation; and \lgf{} is a log-barrier on the fraction of positive
entries, active only when it falls below $\gamma$. None of the three
depends on the conditioning variable, so they regulate signal scale
rather than altering the learned partition.

$T_\theta$ is a residual MLP with a two-layer input projection
($2d{\to}256{\to}256$, $d{=}8$), a learned position embedding of width
$64$, three residual blocks of width $512$, and a linear head to
$V{=}16\,384$; it has $9.44$M parameters, $8.40$M of them in the head.
It is trained on $5\,000$ clean token sequences from the frozen
generator on ImageNet with AdamW ($\mathrm{lr}=10^{-4}$,
$\beta=(0.9,0.95)$), batch size $8$, for $10\,000$ steps in
mixed-precision bf16 on a single A100-40G. The attack layer samples
with probabilities
$\{$identity$:0.5$, JPEG $q{\in}[10,30]:0.2$, Gaussian noise
$\sigma{\in}[0.01,0.05]:0.15$, Gaussian blur $k{\in}[3,7]:0.15\}$.
\emph{Baselines.} ClusterMark~\cite{lukovnikov2026clustermark} uses
$k$-means clustering with $k{=}64$ at $\gamma{=}0.25$;
IndexMark~\cite{tong2025indexmark} uses training-free MWPM pairing at
$\gamma{=}0.5$, reported at both $\tau{=}0$ and the paper-recommended
$\tau{=}0.8$.

\emph{Calibrating $\delta$ across binary and continuous rules.}
The hash baselines and \method{} differ in the range of
$\mathbf{w}_t$ in Eq.~(2), so equal $\delta$ would not give an
equal logit shift. The hash baselines use the standard KGW rule, adding
the bias to greenlist logits only
($w_t(z)\!\in\!\{0,1\}$, implemented as
\texttt{logits}${+}\delta\cdot\mathbb{1}[z\text{ green}]$), so their
green-to-red logit gap is exactly $\delta$. \method{} adds
$\delta\,w_t(z)$ with $w_t(z)\!\in\!({-}1,{+}1)$ from a $\tanh$, so its
gap is $\delta\,(w_t^{\text{green}}{-}w_t^{\text{red}})$, which attains
$2\delta$ only where the $\tanh$ saturates and is smaller everywhere
else. Running the baselines at $\delta{=}8$ against our $\delta{=}4$
therefore equalizes the maximum gap at $8$ logits; because our $\tanh$
is rarely saturated at $k_2{=}2$, our realized gap stays below that
ceiling while the baselines sit at theirs, making the comparison
conservative in the baselines' favor.

\emph{Losses.} The grouped terms of Eq.~(5) expand as
$\mathcal{L}_\mathrm{wm}{=}\ldet{+}\mathcal{L}_q$ (both unweighted),
$\mathcal{L}_\mathrm{sm}{=}\mathcal{L}_s{+}\mathcal{L}_\mathrm{cb}$
(equal weights, hence the shared $\lambda_\mathrm{sm}{=}0.1$), and
$\mathcal{L}_\mathrm{cal}$ as above. Coefficients are
$\lambda_\mathrm{sm}{=}0.1$, $\lambda_\mathrm{pc}{=}2.5$, and
$(\lambda_n,\lambda_\mathrm{gf},\lambda_\mathrm{ent})
=(0.01,10.0,0.5)$. We use $k_1{=}20$, $k_2{=}2$,
$\mathrm{top}K{=}100$ and $n_\mathrm{pairs}{=}16$ in \lpc{}, and $8$
nearest neighbors in $\mathcal{L}_\mathrm{cb}$. Training on LlamaGen-B
takes four hours; on LlamaGen-L it takes five. All reported FID numbers
use the standard torchmetrics implementation with Inception-V3 pool3
features against a size-matched clean partition drawn with the same
sampling parameters as the watermarked set.

In deployment, each model provider trains and retains its own private
signal network $T_\theta$ alongside secret keys in accordance with the
threat model of Sec.~3. Training, evaluation, and configuration
code to reproduce all experiments in this paper will be released upon
publication.

\section{Post-Edit Detection and Image-Attack Robustness}
\label{sec:supp_edit_rob}

\subsection{Semantic-preservation screening of the edit suite}
\label{supp:semprep}

The eleven edits are: JPEG and WebP at $q{=}25$ (plus $q{=}50$ for the
consistency analysis); a brightness shift of $0.4$; a saturation shift
of $0.6$; a combined color jitter (brightness $0.3$, contrast $0.3$,
hue $0.15$, saturation $0.5$); Gaussian noise $\sigma{=}0.05$; Gaussian
blur $r{=}2$; a $50\%$ downscale-and-restore; a $5^\circ$ rotation; and
a $5\%$ random crop.

To verify that the edit suite preserves image semantics (screening all ten
distinct transforms at their harshest quality setting, since JPEG $q{=}25$
bounds $q{=}50$), we score each original-to-edited pair using encoders
independent of watermarks and tokens: cosine similarity of DINOv2 ViT-L/14
CLS features~\cite{oquab2024dinov2} (instance-level identity), cosine
similarity of CLIP ViT-L/14 image embeddings~\cite{radford2021clip}
(caption-level semantics), and LPIPS distance~\cite{zhang2018lpips},
measured over $200$ watermarked LlamaGen-B images (\cref{tab:semprep}).
Two reference points calibrate the scale: the \emph{cross-source} row
pairs two unrelated generated images (DINOv2 cosine $0.030$, LPIPS
$0.835$), while the SD img2img rows mark content-altering regeneration.

Every edit in the suite stays at DINOv2 cosine ${\geq}0.870$ and CLIP
cosine ${\geq}0.841$, far above the cross-source floor, whereas SD
img2img at $s{=}0.5$ falls to DINOv2 $0.578$ (below every edit in the
suite), motivating our treatment of regeneration as a separate attack
family (Tab.~3). Moreover, token drift and semantic change are
decoupled: a $5^\circ$ rotation rewrites $95.7\%$ of the tokens
(Tab.~2) yet preserves content best among the harder
edits (DINOv2 $0.934$), whereas WebP~$q25$ rewrites $63.2\%$ and has the
lowest DINOv2 similarity ($0.870$). Pixel PSNR is similarly sensitive to
spatial shift ($16.3$ and $19.9$\,dB on rotation and crop despite high
semantic similarity), showing that an image can remain visually
recognizable while becoming an almost completely rewritten token sequence.

\begin{table}[!htb]
\centering
\scriptsize
\setlength{\tabcolsep}{3.5pt}
\caption{\textbf{Semantic-preservation screening} of the edit suite
(LlamaGen-B, $200$ watermarked images, original vs.\ edited), sorted by
DINOv2 similarity. \emph{cross-source} pairs two unrelated images as the
content floor; SD img2img rows calibrate content-altering regeneration;
``drift'' repeats the token change rate from Tab.~2.}
\label{tab:semprep}
\begin{tabular}{@{}lcccc|c@{}}
\toprule
Edit & DINOv2$\uparrow$ & CLIP$\uparrow$ & LPIPS$\downarrow$ & PSNR & drift \\
\midrule
saturation            & 0.997 & 0.991 & 0.018 & 37.6 & 0.100 \\
color jitter          & 0.979 & 0.949 & 0.125 & 21.3 & 0.539 \\
brightness            & 0.976 & 0.983 & 0.037 & 23.8 & 0.251 \\
crop $5\%$            & 0.958 & 0.972 & 0.105 & 19.9 & 0.854 \\
blur $r{=}2$          & 0.957 & 0.929 & 0.178 & 32.3 & 0.410 \\
resize $50\%$         & 0.949 & 0.939 & 0.113 & 33.4 & 0.357 \\
rotate $5^\circ$      & 0.934 & 0.943 & 0.187 & 16.3 & 0.957 \\
noise $\sigma{=}0.05$ & 0.904 & 0.942 & 0.216 & 26.6 & 0.544 \\
JPEG $q25$            & 0.897 & 0.875 & 0.098 & 31.4 & 0.607 \\
WebP $q25$            & 0.870 & 0.841 & 0.090 & 31.5 & 0.632 \\
\midrule
\multicolumn{6}{@{}l}{\itshape Regeneration, for calibration (not in the edit suite)} \\
SD img2img $s{=}0.15$ & 0.848 & 0.939 & 0.093 & 26.2 & --- \\
SD img2img $s{=}0.30$ & 0.761 & 0.916 & 0.155 & 23.6 & --- \\
SD img2img $s{=}0.50$ & 0.578 & 0.870 & 0.271 & 20.8 & --- \\
\midrule
\textit{cross-source floor} & 0.030 & 0.669 & 0.835 & 7.4 & 1.000 \\
\bottomrule
\end{tabular}
\end{table}

\subsection{Paired bootstrap confidence intervals}
\label{supp:ci}

To verify the statistical reliability of the AUROC margins in
Tab.~1 on $400$ test images, \cref{tab:edit_ci} reports
$95\%$ paired bootstrap confidence intervals over $2\,000$ resampled
replicates, scoring identical image pairs across schemes in every
replicate.

Against one-token KGW ($h{=}1$), \method{} achieves statistically
significant gains on eight of the ten edits, including decisive margins on
$5^\circ$ rotation ($[+0.145,+0.244]$) and $5\%$ cropping
($[+0.129,+0.191]$). Against global KGW ($h{=}0$), per-edit differences
stay close to zero across nearly all edits, confirming that \method{}
matches the edit robustness of a static global partition while avoiding
its catastrophic secrecy failure ($0.943$ vs.\ $0.015$ post-removal AUROC).

\begin{table}[!htb]
\centering
\footnotesize
\caption{Paired bootstrap ($2\,000$ replicates, $95\%$ intervals) on
the AUROC differences of Tab.~1, LlamaGen-B. A
starred interval excludes zero; differences are computed at full
precision before rounding.}
\label{tab:edit_ci}
\setlength{\tabcolsep}{2.0pt}
\scriptsize
\begin{tabular}{@{}l|c|cc@{}}
\toprule
Edit & \method{} AUROC & $\Delta$ vs.\ $h{=}0$ & $\Delta$ vs.\ $h{=}1$ \\
\midrule
clean        & 1.000 \tiny[1.00,1.00] & $+.000$ \tiny[$-$.000,$+$.000] & $-.000$ \tiny[$-$.000,$+$.000] \\
JPEG $q25$   & 0.988 \tiny[.978,.995] & $-.005$ \tiny[$-$.015,$+$.005] & $+.011$ \tiny[$+$.000,$+$.022]$^\star$ \\
WebP $q25$   & 0.981 \tiny[.971,.990] & $-.014$ \tiny[$-$.025,$-$.004]$^\star$ & $+.021$ \tiny[$+$.006,$+$.036]$^\star$ \\
jitter       & 0.989 \tiny[.979,.996] & $-.006$ \tiny[$-$.016,$+$.003] & $+.009$ \tiny[$-$.002,$+$.020] \\
brightness   & 0.998 \tiny[.993,1.00] & $-.002$ \tiny[$-$.007,$+$.001] & $-.002$ \tiny[$-$.007,$+$.000] \\
saturation   & 1.000 \tiny[1.00,1.00] & $-.000$ \tiny[$-$.000,$+$.000] & $+.000$ \tiny[$-$.000,$+$.000] \\
noise        & 0.993 \tiny[.988,.997] & $-.004$ \tiny[$-$.010,$+$.002] & $+.007$ \tiny[$+$.001,$+$.015]$^\star$ \\
blur $r{=}2$ & 1.000 \tiny[.999,1.00] & $+.002$ \tiny[$-$.001,$+$.006] & $+.003$ \tiny[$+$.000,$+$.006]$^\star$ \\
resize $50\%$& 1.000 \tiny[1.00,1.00] & $+.000$ \tiny[$-$.000,$+$.000] & $+.002$ \tiny[$+$.000,$+$.005]$^\star$ \\
\midrule
rot $5^\circ$& 0.822 \tiny[.793,.850] & $+.011$ \tiny[$-$.030,$+$.051] & $+.193$ \tiny[$+$.145,$+$.244]$^\star$ \\
crop $5\%$   & 0.961 \tiny[.949,.973] & $+.004$ \tiny[$-$.011,$+$.021] & $+.158$ \tiny[$+$.129,$+$.191]$^\star$ \\
\bottomrule
\end{tabular}
\end{table}

\subsection{LlamaGen-B block of the main robustness table}
\label{supp:geomB}

Tab.~3 reports LlamaGen-L for space, and \cref{tab:geom_B}
provides the LlamaGen-B counterpart over the identical attack set and
markings for direct row-by-row comparison. All image-level rows use
$\delta{=}4$ at both scales, while regeneration rows use $\delta{=}8$ on
LlamaGen-B (matching the stronger regeneration perturbation regime on the
smaller $111$M generator) and $\delta{=}4$ on LlamaGen-L (Tab.~3).
Across scales, the ordering stays consistent: \method{} matches or
outperforms the best token-level baseline on nine of the fifteen rows on
LlamaGen-B and on eleven of the fifteen on LlamaGen-L (Tab.~3),
leads all methods on diffusion regeneration, and trades off VQ-$k$NN
remapping against ClusterMark's cluster-level coloring.

\begin{table}[!htb]
\centering
\caption{\textbf{Robustness on LlamaGen-B} (AUROC), the counterpart of
Tab.~3. Markings as in Tab.~3: bold$+$underlined is
the best in the block, \textcolor{lossRed}{$\bigstar$} marks rows where
\method{} matches or exceeds the best token-level baseline, and the
parenthesized value is the signed margin against it.}
\label{tab:geom_B}
\scriptsize
\setlength{\tabcolsep}{2.5pt}
\setlength{\cnumwd}{2.3em}\setlength{\cstarwd}{0.7em}\setlength{\cannwd}{2.6em}
\begin{tabular}{@{}l|cccc@{}}
\toprule
Attack & \method{} & KGW $h{=}0$ & CM & IM \\
\midrule
clean                        & \cneu{1.000} & \best{1.000} & \best{1.000} & \best{1.000} \\
\midrule
\multicolumn{5}{@{}l}{\itshape Geometric transformations} \\
\quad rot $5^\circ$          & \cwin{0.822}{\gain{0.011}} & 0.811 & 0.799 & 0.640 \\
\quad rot $30^\circ$         & \clos{0.592}{\loss{0.012}} & \textbf{\underline{0.604}} & 0.587 & 0.531 \\
\quad rot $90^\circ$         & \cwin{0.586}{\gain{0.004}} & 0.550 & 0.582 & 0.527 \\
\quad hflip                  & \cwin{0.742}{\gain{0.006}} & 0.736 & 0.648 & 0.627 \\
\quad vflip                  & \clos{0.605}{\loss{0.084}} & \textbf{\underline{0.689}} & 0.665 & 0.586 \\
\quad resize $25\%$          & \cwin{0.970}{\gain{0.003}} & 0.951 & 0.967 & 0.834 \\
\midrule
\multicolumn{5}{@{}l}{\itshape Compression, noise \& blur} \\
\quad JPEG $q50$             & \clos{0.999}{\loss{0.001}} & 0.999 & \textbf{\underline{1.000}} & 0.991 \\
\quad noise $\sigma{=}.05$   & \clos{0.992}{\loss{0.005}} & \textbf{\underline{0.997}} & 0.994 & 0.966 \\
\quad blur $r{=}2$           & \cwin{1.000}{\gain{0.000}} & 0.998 & 1.000 & --- \\
\midrule
\multicolumn{5}{@{}l}{\itshape Combined pipelines} \\
\quad hflip$+$JPEG $q25$     & \cwin{0.589}{\gain{0.018}} & 0.571 & 0.547 & 0.571 \\
\quad rot $5^\circ$$+$crop $5\%$ & \cwin{0.857}{\gain{0.005}} & 0.852 & 0.814 & 0.643 \\
\quad resize $50\%$$+$noise  & \clos{0.984}{\loss{0.012}} & \textbf{\underline{0.996}} & 0.993 & --- \\
\midrule
\multicolumn{5}{@{}l}{\itshape Regeneration} \\
\quad VQ regen ($k{=}5$)     & \clos{0.852}{\loss{0.112}} & 0.613 & \textbf{\underline{0.964}} & 0.454 \\
\quad Diff.\ regen ($s{=}.5$) & \cwin{0.696}{\gain{0.105}} & 0.574 & 0.591 & 0.533 \\
\bottomrule
\end{tabular}
\end{table}

\subsection{Full image-attack robustness table}
\label{sec:supp_geom}

Tab.~3 reports a representative subset of attacks.
\cref{tab:geom_full} presents all twenty-one image-level attacks across
both generator scales, adding additional rescale, compression, noise, and
blur strengths from the standard image-watermarking benchmark. In
addition, increasing $\delta$ from $4$ to $8$ on LlamaGen-B consistently
boosts geometric and combined-pipeline AUROC (e.g., $+0.036$ on $5^\circ$
rotation, $+0.055$ on $30^\circ$ rotation, and $+0.031$ on
hflip$+$JPEG~$q25$), confirming that stronger signal amplitude primarily
benefits spatial rearrangement settings while $\delta{=}4$ already
saturates lossy requantization.

\begin{table*}[!t]
\centering
\footnotesize
\setlength{\tabcolsep}{4pt}
\caption{\textbf{AUROC under the full twenty-one-attack image-level
suite on LlamaGen-B and LlamaGen-L} ($\delta{=}4$ for \method{},
$\tau{=}0$ for IndexMark [IM], and matched baseline settings of
Sec.~4.1 and \cref{supp:implementation}). Representative rows appear
in~Tab.~3. Bold: best within each generator's $4$-column
block; cells marked ``---'' were not re-run for IM on that generator.
Nine additional strengths in the rescale, compression, noise, and blur
families complement the main-paper subset across the standard
image-watermarking benchmark~\cite{tancik2020stegastamp,wen2023treering}.}
\label{tab:geom_full}
\begin{tabular}{@{}l cccc cccc@{}}
\toprule
 & \multicolumn{4}{c}{\textbf{LlamaGen-B} (111\,M)}
 & \multicolumn{4}{c}{\textbf{LlamaGen-L} (343\,M)} \\
\cmidrule(lr){2-5}\cmidrule(lr){6-9}
Attack & \method{} & KGW $h{=}0$ & CM & IM & \method{} & KGW $h{=}0$ & CM & IM \\
\midrule
clean                                  & \best{1.000} & \best{1.000} & \best{1.000} & \best{1.000} & \best{1.000} & \best{1.000} & \best{1.000} & \best{1.000} \\
\multicolumn{9}{@{}l}{\itshape Geometric} \\
rotate $5^\circ$                       & \best{0.822} & 0.811 & 0.799 & 0.640 & \best{0.807} & 0.805 & 0.791 & 0.622 \\
rotate $30^\circ$                      & 0.592 & \best{0.604} & 0.587 & 0.531 & \best{0.612} & 0.606 & 0.608 & 0.582 \\
rotate $90^\circ$                      & \best{0.586} & 0.550 & 0.582 & 0.527 & \best{0.582} & 0.579 & 0.580 & 0.518 \\
hflip                                  & \best{0.742} & 0.736 & 0.648 & 0.627 & \best{0.763} & 0.725 & 0.646 & 0.573 \\
vflip                                  & 0.605 & \best{0.689} & 0.665 & 0.586 & \best{0.665} & 0.659 & 0.643 & 0.566 \\
\multicolumn{9}{@{}l}{\itshape Rescale} \\
resize $75\%$                          & \best{1.000} & 1.000 & 1.000 & 1.000 & \best{1.000} & 1.000 & 1.000 & 0.997 \\
resize $50\%$                          & \best{1.000} & 1.000 & 1.000 & 0.995 & \best{1.000} & 1.000 & 1.000 & 0.992 \\
resize $25\%$                          & \best{0.970} & 0.951 & 0.967 & 0.834 & 0.954 & 0.955 & \best{0.961} & 0.792 \\
\multicolumn{9}{@{}l}{\itshape Compression} \\
JPEG $q25$                             & 0.988 & \best{0.993} & 0.992 & 0.968 & 0.991 & \best{0.994} & 0.993 & 0.954 \\
JPEG $q50$                             & 0.999 & 0.999 & \best{1.000} & 0.991 & \best{1.000} & 0.998 & 1.000 & 0.987 \\
JPEG $q75$                             & 0.999 & \best{1.000} & 1.000 & 0.997 & \best{1.000} & 0.999 & 1.000 & --- \\
WebP $q25$                             & 0.981 & \best{0.995} & 0.994 & 0.967 & 0.985 & 0.993 & \best{0.996} & 0.956 \\
WebP $q50$                             & 0.997 & 0.998 & \best{1.000} & --- & 0.996 & \best{0.999} & 0.998 & --- \\
WebP $q75$                             & 0.998 & \best{0.999} & 0.999 & 0.995 & 0.998 & 0.997 & \best{0.999} & 0.993 \\
\multicolumn{9}{@{}l}{\itshape Noise and blur} \\
Gauss.\ noise $\sigma{=}0.05$          & 0.992 & \best{0.997} & 0.994 & 0.966 & \best{0.997} & 0.993 & 0.991 & 0.953 \\
Gauss.\ noise $\sigma{=}0.10$          & 0.944 & \best{0.984} & 0.975 & 0.900 & 0.970 & \best{0.975} & 0.960 & --- \\
Gauss.\ blur $r{=}2$                   & \best{1.000} & 0.998 & 1.000 & --- & \best{1.000} & 1.000 & 1.000 & --- \\
Gauss.\ blur $r{=}4$                   & 0.986 & 0.990 & \best{0.991} & 0.927 & 0.972 & \best{0.989} & 0.981 & --- \\
\multicolumn{9}{@{}l}{\itshape Combined} \\
rot$5^\circ+$crop$5\%$                 & \best{0.857} & 0.852 & 0.814 & 0.643 & 0.815 & \best{0.874} & 0.808 & 0.648 \\
hflip$+$JPEG $q25$                     & \best{0.589} & 0.571 & 0.547 & 0.571 & \best{0.596} & 0.589 & 0.558 & 0.500 \\
resize $50\%+$noise $\sigma{=}0.05$    & 0.984 & \best{0.996} & 0.993 & --- & \best{0.993} & 0.989 & 0.990 & --- \\
\bottomrule
\end{tabular}
\end{table*}

\paragraph{Pixel-domain reference.}
Tab.~3 includes DWT-DCT~\cite{khan2013digitalwatermarkingimageauthenticationbased}
as a generator-agnostic point of comparison. \cref{tab:dwt_ref} lists its
AUROC across all evaluated image-level and regeneration attacks on
LlamaGen-B. DWT-DCT is competitive only where the pixel grid survives
intact, such as vertical flip ($0.888$) and resize~$50\%{+}$noise
($0.860$), and falls near chance under compression ($0.528$ to $0.569$).
Crucially for Sec.~4.2.4, all three regeneration attacks drive
DWT-DCT to chance (auto-encoder $0.524$, VQ-$k$NN $0.525$, diffusion
$0.550$), because resynthesizing the image from a latent discards the
pixel-frequency carrier outright, whereas token-level schemes retain a
strong detection margin (Tab.~3, \cref{tab:geom_B}).

\begin{table}[!htb]
\centering
\scriptsize
\caption{\textbf{DWT-DCT pixel-domain reference on LlamaGen-B} (AUROC
against unwatermarked negatives; clean detection is $0.893$), covering
both image-level transforms and the three regeneration attacks outside
\cref{tab:geom_full}.}
\label{tab:dwt_ref}
\setlength{\tabcolsep}{3.0pt}
\begin{tabular}{@{}lc@{\hspace{8pt}}lc@{}}
\toprule
Attack & AUROC & Attack & AUROC \\
\midrule
\multicolumn{2}{@{}l}{\itshape Geometric} & \multicolumn{2}{l}{\itshape Compression} \\
rotate $5^\circ$          & 0.741 & JPEG $q25$             & 0.528 \\
rotate $30^\circ$         & 0.562 & JPEG $q50$             & 0.532 \\
rotate $90^\circ$         & 0.331 & JPEG $q60$             & 0.569 \\
hflip                     & 0.460 & \multicolumn{2}{l}{\itshape Noise and blur} \\
vflip                     & 0.888 & noise $\sigma{=}0.05$  & 0.862 \\
\multicolumn{2}{@{}l}{\itshape Rescale and crop} & noise $\sigma{=}0.20$ & 0.456 \\
resize $25\%$             & 0.610 & salt\&pepper $10\%$    & 0.795 \\
crop $5\%$                & 0.803 & blur $r{=}2$           & 0.752 \\
crop $10\%$               & 0.714 & blur $r{=}3$           & 0.623 \\
center crop $25\%$        & 0.458 & color jitter           & 0.659 \\
\multicolumn{2}{@{}l}{\itshape Combined} & \multicolumn{2}{l}{\itshape Regeneration} \\
rot $5^\circ{+}$crop $5\%$ & 0.731 & auto-encoder          & 0.524 \\
hflip${+}$JPEG $q25$       & 0.572 & VQ-$k$NN ($k{=}5$)    & 0.525 \\
resize $50\%{+}$noise      & 0.860 & diffusion ($s{=}0.5$) & 0.550 \\
\bottomrule
\end{tabular}
\end{table}

\subsection{Robustness panorama across attack strengths}
\label{supp:panorama}

\begin{figure*}[!t]
\centering
\includegraphics[width=\textwidth]{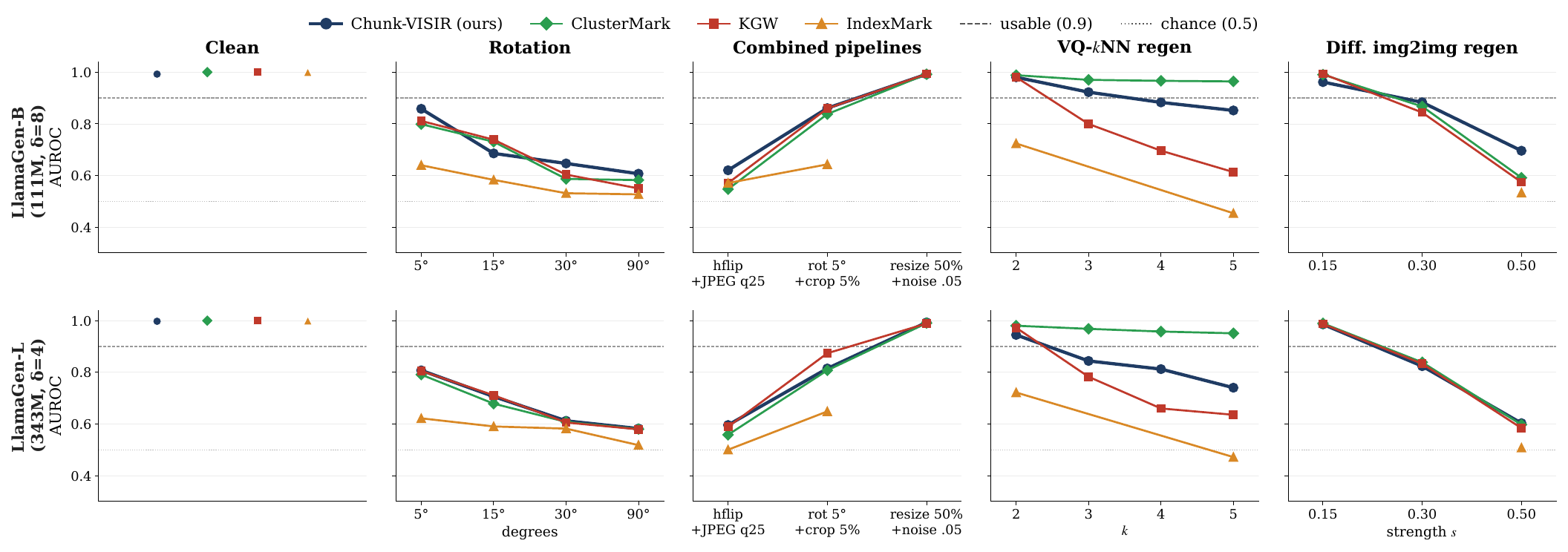}
\caption{\textbf{Robustness panorama across 5 attack families and 2
generator scales} (detection AUROC), expanding the single-strength rows
of~Tab.~3 into full sweeps. Rows: LlamaGen-B ($\delta{=}8$,
matching the~Tab.~3 regeneration operating point) and L
($\delta{=}4$). Columns: Clean / Rotation $\{5,15,30,90\}^\circ$ /
three composite real-world pipelines (hflip$+$JPEG $q25$, rot
$5^\circ+$crop $5\%$, resize $50\%+$noise $\sigma{=}.05$) / VQ-$k$NN
regeneration $k\!\in\!\{2,3,4,5\}$ / SD img2img
$s\!\in\!\{0.15,0.30,0.50\}$. Dashed line: $0.9$ usability threshold;
dotted: chance ($0.5$).}
\label{fig:panorama}
\end{figure*}

To confirm that the method ordering in Tab.~3 holds across full
perturbation sweeps rather than only at fixed single-strength operating
points, \cref{fig:panorama} sweeps five attack families across both
generator scales. The relative ordering remains consistent along every
curve: wherever \method{} leads at the canonical strength in
Tab.~3, it maintains that advantage across the full sweep, while
ClusterMark's advantage on VQ-$k$NN remapping reflects its discrete
cluster-level coloring across all $k$.

\subsection{Qualitative robustness panel}
\label{supp:qualitative}

\begin{figure*}[!tb]
\centering
\setlength{\tabcolsep}{1pt}
\renewcommand{\arraystretch}{0.95}
\footnotesize
\begin{tabular}{@{}ccccccc@{}}
clean & WM & \scriptsize +rot $5^\circ$ & \scriptsize +hflip
      & \scriptsize +JPEG $q25$ & \scriptsize +resize $50\%$
      & \scriptsize +noise $\sigma{=}.05$ \\[1pt]
\includegraphics[width=.136\textwidth]{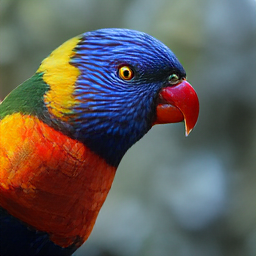} &
\includegraphics[width=.136\textwidth]{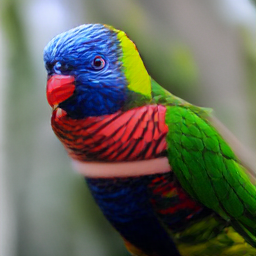} &
\includegraphics[width=.136\textwidth]{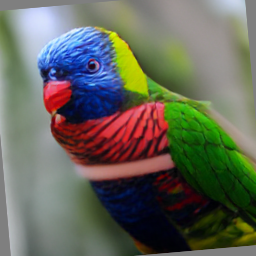} &
\includegraphics[width=.136\textwidth]{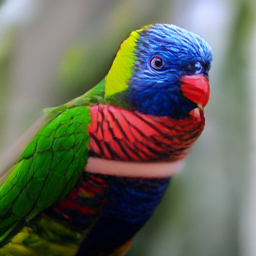} &
\includegraphics[width=.136\textwidth]{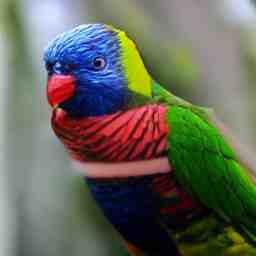} &
\includegraphics[width=.136\textwidth]{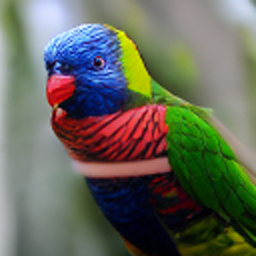} &
\includegraphics[width=.136\textwidth]{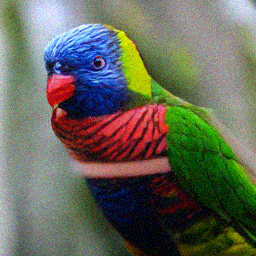} \\[-2pt]
\multicolumn{7}{@{}l}{\scriptsize lorikeet} \\[2pt]
\includegraphics[width=.136\textwidth]{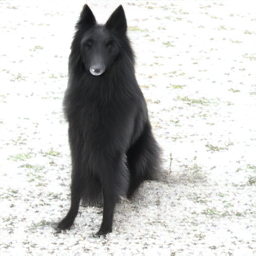} &
\includegraphics[width=.136\textwidth]{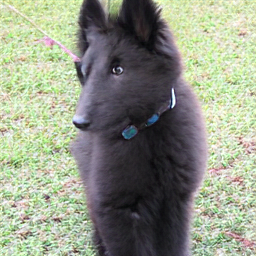} &
\includegraphics[width=.136\textwidth]{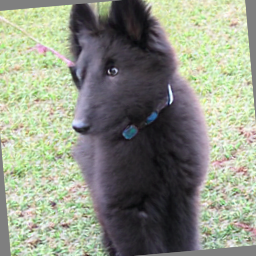} &
\includegraphics[width=.136\textwidth]{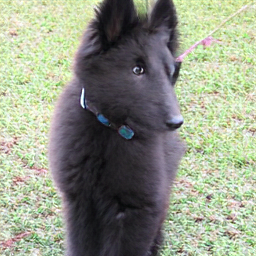} &
\includegraphics[width=.136\textwidth]{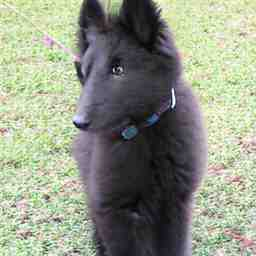} &
\includegraphics[width=.136\textwidth]{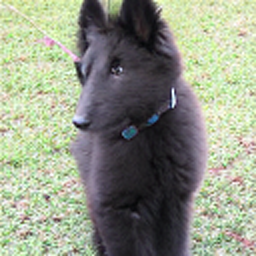} &
\includegraphics[width=.136\textwidth]{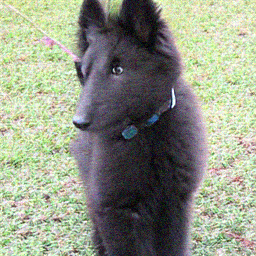} \\[-2pt]
\multicolumn{7}{@{}l}{\scriptsize groenendael} \\[2pt]
\includegraphics[width=.136\textwidth]{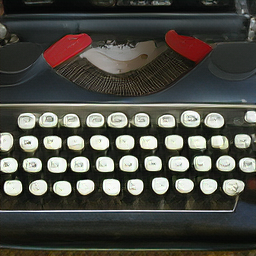} &
\includegraphics[width=.136\textwidth]{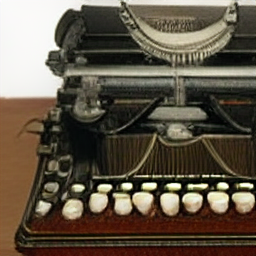} &
\includegraphics[width=.136\textwidth]{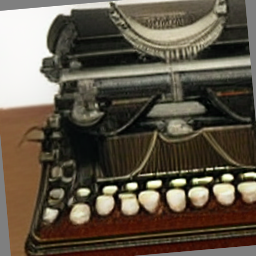} &
\includegraphics[width=.136\textwidth]{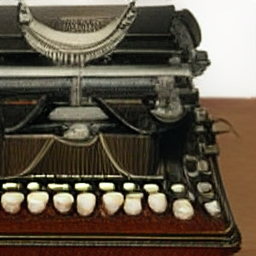} &
\includegraphics[width=.136\textwidth]{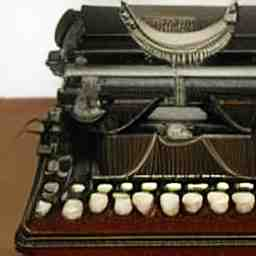} &
\includegraphics[width=.136\textwidth]{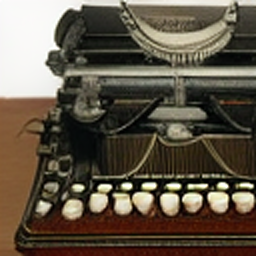} &
\includegraphics[width=.136\textwidth]{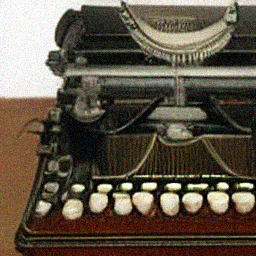} \\[-2pt]
\multicolumn{7}{@{}l}{\scriptsize space bar} \\[2pt]
\includegraphics[width=.136\textwidth]{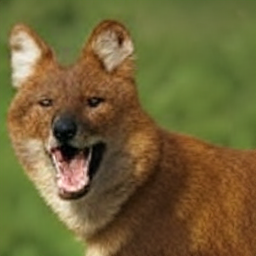} &
\includegraphics[width=.136\textwidth]{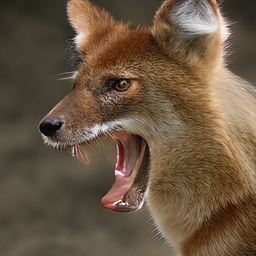} &
\includegraphics[width=.136\textwidth]{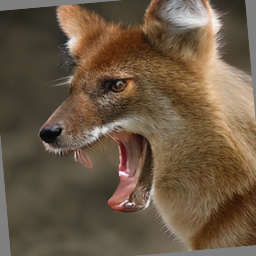} &
\includegraphics[width=.136\textwidth]{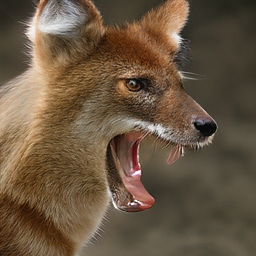} &
\includegraphics[width=.136\textwidth]{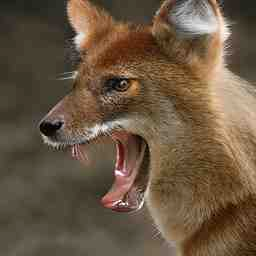} &
\includegraphics[width=.136\textwidth]{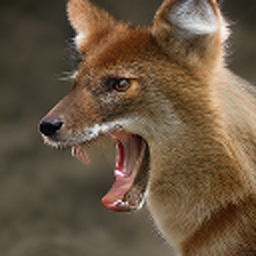} &
\includegraphics[width=.136\textwidth]{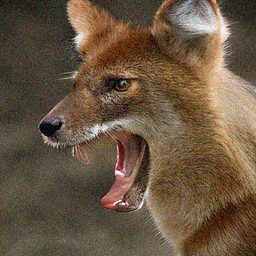} \\[-2pt]
\multicolumn{7}{@{}l}{\scriptsize dhole} \\[2pt]
\includegraphics[width=.136\textwidth]{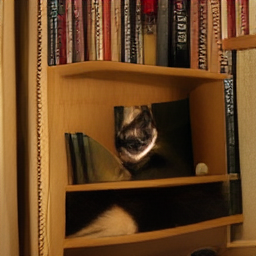} &
\includegraphics[width=.136\textwidth]{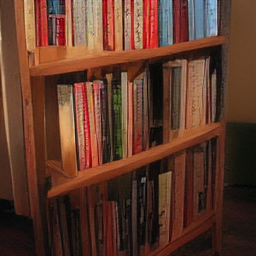} &
\includegraphics[width=.136\textwidth]{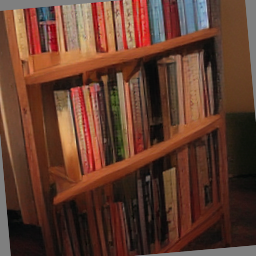} &
\includegraphics[width=.136\textwidth]{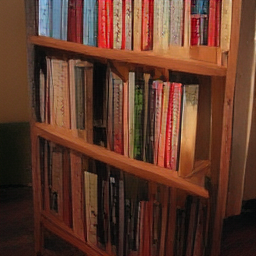} &
\includegraphics[width=.136\textwidth]{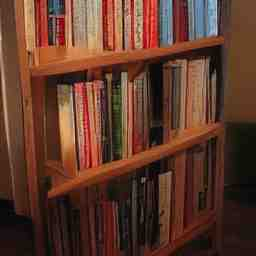} &
\includegraphics[width=.136\textwidth]{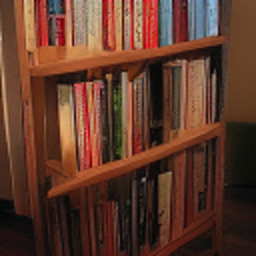} &
\includegraphics[width=.136\textwidth]{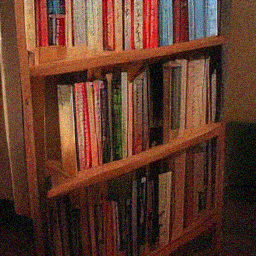} \\[-2pt]
\multicolumn{7}{@{}l}{\scriptsize bookcase} \\[2pt]
\includegraphics[width=.136\textwidth]{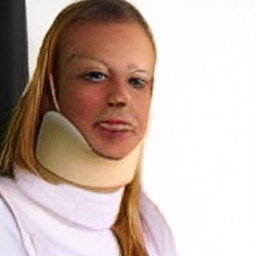} &
\includegraphics[width=.136\textwidth]{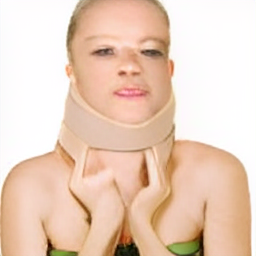} &
\includegraphics[width=.136\textwidth]{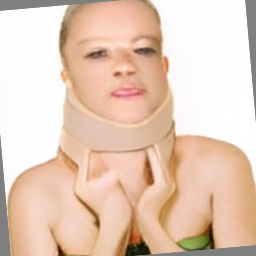} &
\includegraphics[width=.136\textwidth]{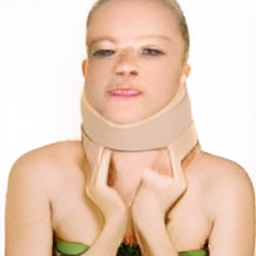} &
\includegraphics[width=.136\textwidth]{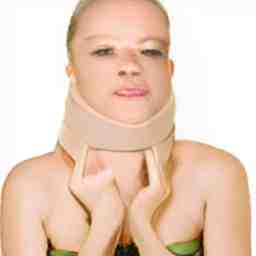} &
\includegraphics[width=.136\textwidth]{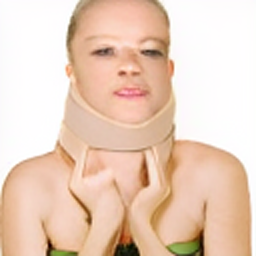} &
\includegraphics[width=.136\textwidth]{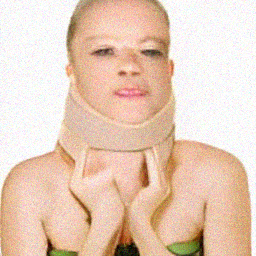} \\[-2pt]
\multicolumn{7}{@{}l}{\scriptsize neck brace} \\
\end{tabular}
\caption{\textbf{Qualitative robustness panel (LlamaGen-L, $\delta{=}4$).}
Six random ImageNet classes (one per row): column~1 is the clean
image, column~2 the \method{}-watermarked image generated with the
same random seed, and columns~3 to~7 apply the five canonical image-domain
attacks to the watermarked image. Class identity, photographic
style, lighting, and texture are preserved end-to-end, and each attack
leaves the detector well above its usability threshold
(Tab.~3).}
\label{fig:appendix_qualitative}
\end{figure*}

\begin{figure*}[!tb]
\centering
\setlength{\tabcolsep}{1pt}
\renewcommand{\arraystretch}{0.95}
\footnotesize
\begin{tabular}{@{}cccc@{\hspace{8pt}}cccc@{}}
clean & \method{} & CM & IM & clean & \method{} & CM & IM \\[1pt]
\includegraphics[width=.117\textwidth]{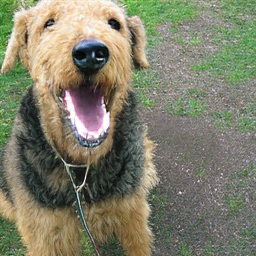} &
\includegraphics[width=.117\textwidth]{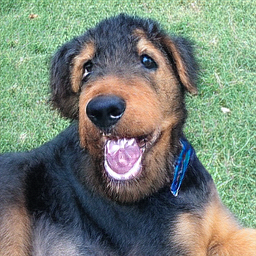} &
\includegraphics[width=.117\textwidth]{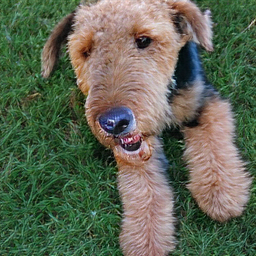} &
\includegraphics[width=.117\textwidth]{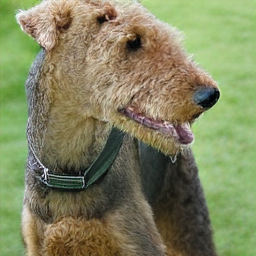} &
\includegraphics[width=.117\textwidth]{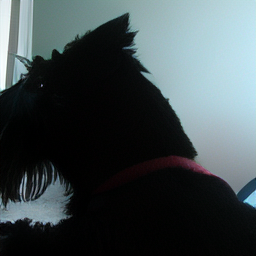} &
\includegraphics[width=.117\textwidth]{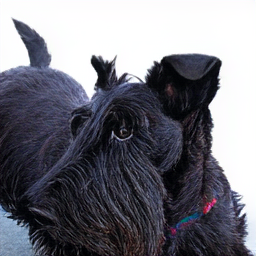} &
\includegraphics[width=.117\textwidth]{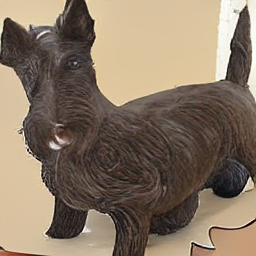} &
\includegraphics[width=.117\textwidth]{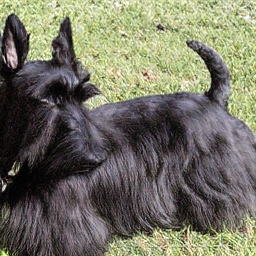} \\[-2pt]
\multicolumn{4}{@{}l}{\scriptsize Airedale} & \multicolumn{4}{@{}l}{\scriptsize Scotch terrier} \\[2pt]
\includegraphics[width=.117\textwidth]{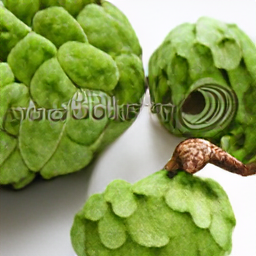} &
\includegraphics[width=.117\textwidth]{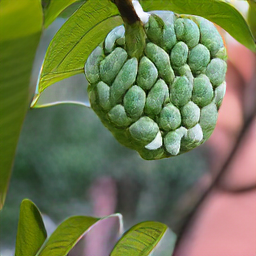} &
\includegraphics[width=.117\textwidth]{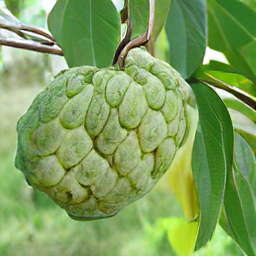} &
\includegraphics[width=.117\textwidth]{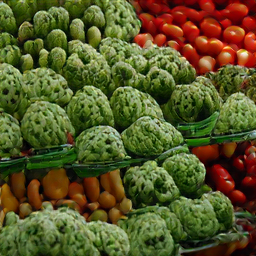} &
\includegraphics[width=.117\textwidth]{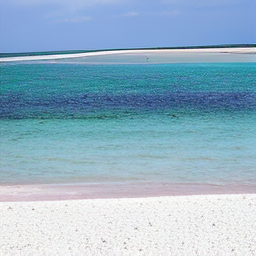} &
\includegraphics[width=.117\textwidth]{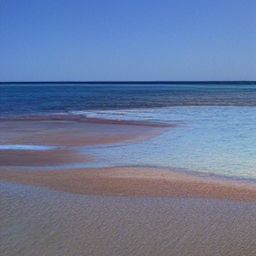} &
\includegraphics[width=.117\textwidth]{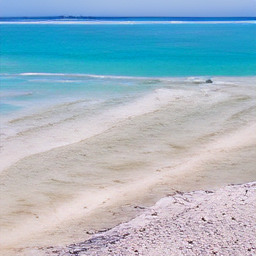} &
\includegraphics[width=.117\textwidth]{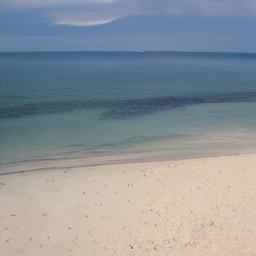} \\[-2pt]
\multicolumn{4}{@{}l}{\scriptsize custard apple} & \multicolumn{4}{@{}l}{\scriptsize sandbar} \\[2pt]
\includegraphics[width=.117\textwidth]{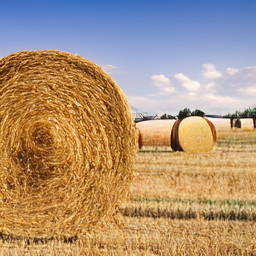} &
\includegraphics[width=.117\textwidth]{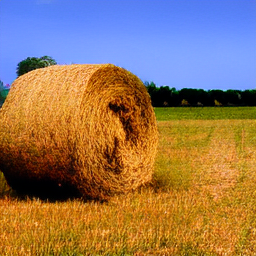} &
\includegraphics[width=.117\textwidth]{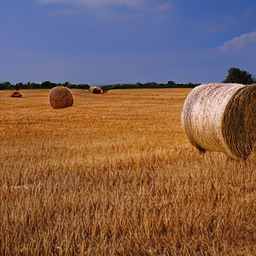} &
\includegraphics[width=.117\textwidth]{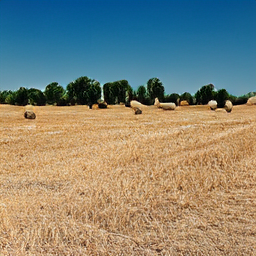} &
\includegraphics[width=.117\textwidth]{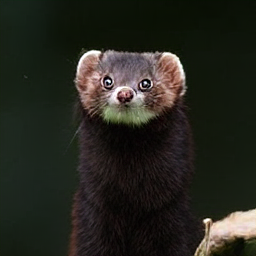} &
\includegraphics[width=.117\textwidth]{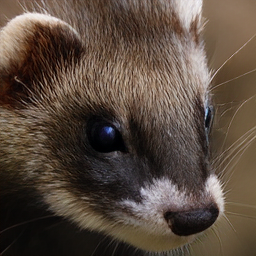} &
\includegraphics[width=.117\textwidth]{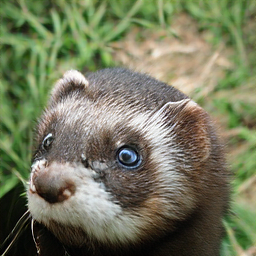} &
\includegraphics[width=.117\textwidth]{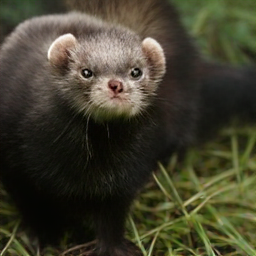} \\[-2pt]
\multicolumn{4}{@{}l}{\scriptsize hay} & \multicolumn{4}{@{}l}{\scriptsize polecat} \\[2pt]
\includegraphics[width=.117\textwidth]{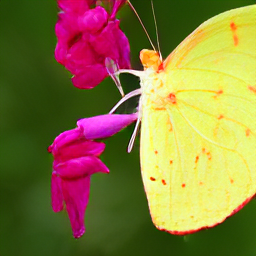} &
\includegraphics[width=.117\textwidth]{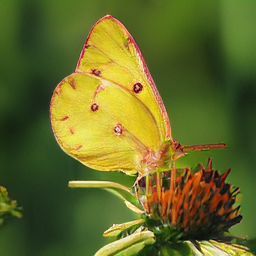} &
\includegraphics[width=.117\textwidth]{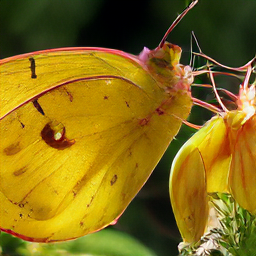} &
\includegraphics[width=.117\textwidth]{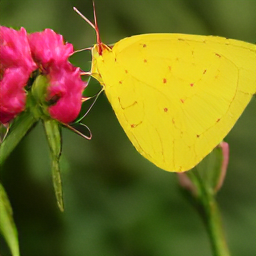} &
\includegraphics[width=.117\textwidth]{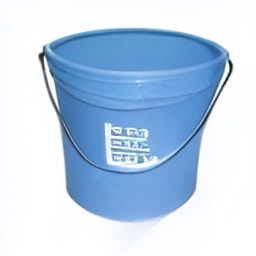} &
\includegraphics[width=.117\textwidth]{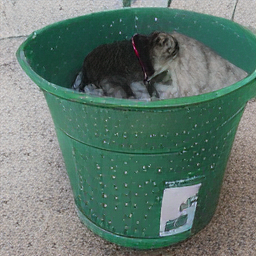} &
\includegraphics[width=.117\textwidth]{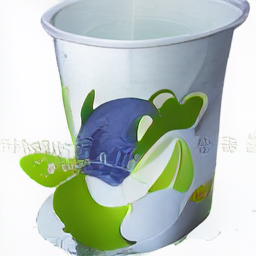} &
\includegraphics[width=.117\textwidth]{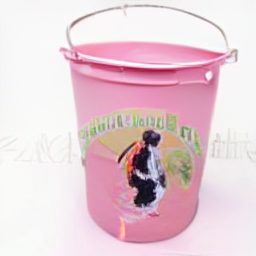} \\[-2pt]
\multicolumn{4}{@{}l}{\scriptsize sulphur butterfly} & \multicolumn{4}{@{}l}{\scriptsize bucket} \\[2pt]
\includegraphics[width=.117\textwidth]{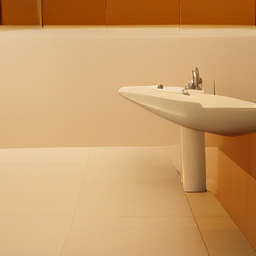} &
\includegraphics[width=.117\textwidth]{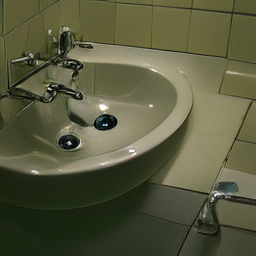} &
\includegraphics[width=.117\textwidth]{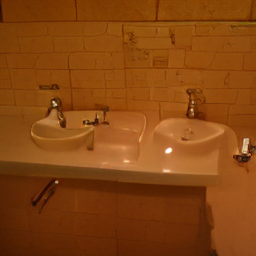} &
\includegraphics[width=.117\textwidth]{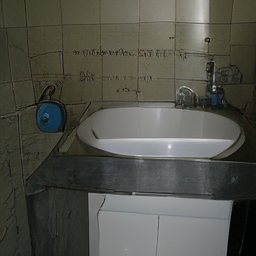} &
\includegraphics[width=.117\textwidth]{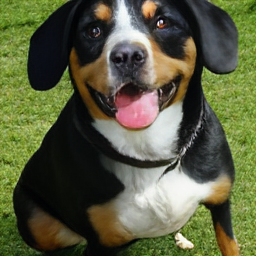} &
\includegraphics[width=.117\textwidth]{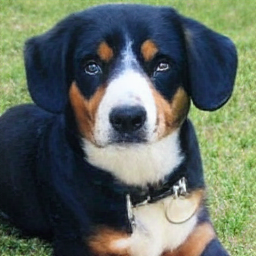} &
\includegraphics[width=.117\textwidth]{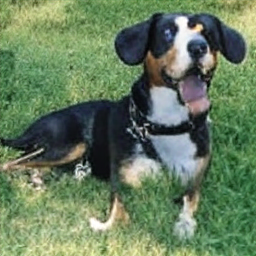} &
\includegraphics[width=.117\textwidth]{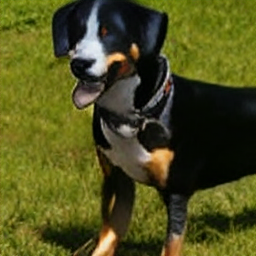} \\[-2pt]
\multicolumn{4}{@{}l}{\scriptsize washbasin} & \multicolumn{4}{@{}l}{\scriptsize EntleBucher} \\[2pt]
\end{tabular}
\caption{\textbf{Same-prompt cross-method comparison (LlamaGen-L,
$256^2$).} Each row shows one ImageNet class generated from the identical
class condition and RNG seed under clean decoding, \method{}
($\delta{=}4$), ClusterMark ($k{=}64$, $\delta{=}5$), and IndexMark
($\tau{=}0$), so cross-column differences stem solely from each scheme's
logit biasing. All token-level watermarks preserve class identity,
composition, and photographic quality, consistent with the narrow FID band
of \cref{tab:fid}.}
\label{fig:fourway_panel}
\end{figure*}

\cref{fig:appendix_qualitative} extends the main-text qualitative
comparison (Fig.~4) to a clean-vs-watermarked-vs-attack
grid. Each row shows a random ImageNet class generated with \method{} on
LlamaGen-L ($\delta{=}4$) using a matched random seed for the clean and
watermarked columns, so any visual difference stems solely from the
watermark. Columns $3$ to $7$ apply the five canonical image-domain
attacks of Tab.~3 (rotate $5^\circ$, hflip, JPEG $q25$, resize
$50\%$, and Gaussian noise $\sigma{=}0.05$); the detector recovers the
watermark above the usability threshold in every case (Tab.~3).
\cref{fig:fourway_panel} additionally compares all token-level schemes on
identical prompts: each row is generated from the same class condition and
RNG seed under clean decoding, \method{}, ClusterMark, and IndexMark at
their canonical operating points, making the columns directly comparable.
Fifty-sample full contact sheets are included with the code release.

\section{Rule Secrecy, Adaptive Attacks, and Multi-Key Isolation}
\label{sec:supp_sec_all}

\subsection{Full removal-attack breakdown (LlamaGen-B and L)}
\label{sec:supp_removal_full}
Fig.~5 in the main paper summarizes the frequency-counting
removal attack. \cref{tab:removal_full} gives the complete LlamaGen-B and -L
breakdown across all baselines, adding $\TPR@1\%$ at both the clean and
post-attack points and the attacker's inferred-vs-true greenlist Jaccard.
The LlamaGen-B operating point ($\delta{=}4$) is the canonical setting used
across all ablations (Sec.~4.4).  Clean columns are measured on
each attack run's own held-out evaluation split rather than the
$400$-image edit benchmark, so \method{}'s LlamaGen-B clean entry reads
$0.996$\,/\,$0.975$ here versus $1.000$\,/\,$1.000$ in
Tab.~1; the same convention applies to
\cref{tab:ablation_removal,tab:scaling}.

\begin{table*}[!t]
\centering
\footnotesize
\setlength{\tabcolsep}{4.5pt}
\caption{\textbf{Adaptive removal attack, full breakdown.}
Each attacker queries $N_\mathrm{atk}\in\{300,500\}$ watermarked images
and is scored on a disjoint held-out split of $300$ to $400$ images;
\cref{tab:natk_full} shows post-removal AUROC is flat in
$N_\mathrm{atk}$ across a $50\times$ range for each scheme it covers, so
these budgets are directly comparable.
$\delta$: KGW $h{=}0$/\method{}/CM bias; $\tau$: IM confidence.
\textbf{Retention}
$=\mathrm{AUROC}_{\text{rem}}/\mathrm{AUROC}_{\text{clean}}$, shaded
\colorbox{green!16}{green}/\colorbox{red!12}{red} for survives/removed.
``Jacc.'' is the attacker's inferred-vs-true greenlist overlap
(defined only for single-split schemes).}
\label{tab:removal_full}
\begin{tabular}{@{}ll cc cc c c c@{}}
\toprule
\multirow{2}{*}{Model} & \multirow{2}{*}{Method (op)}
       & \multicolumn{2}{c}{Clean} & \multicolumn{2}{c}{After removal}
       & \multirow{2}{*}{\makecell{Reten-\\tion}}
       & \multirow{2}{*}{\makecell{Att.\\Jacc.}}
       & \multirow{2}{*}{\makecell{Still\\usable?}} \\
\cmidrule(lr){3-4}\cmidrule(lr){5-6}
       &  & AUROC & $\TPR@1\%$ & AUROC & $\TPR@1\%$ & & & \\
\midrule
\multirow{8}{*}{LlamaGen-B}
      & KGW $h{=}0$ ($\delta{=}4$)          & 1.000 & 1.000 & 0.015 & 0.007 & \rbrk{$1.5\%$}  & 0.959 & \bad \\
      & KGW $h{=}0$ ($\delta{=}8$)          & 1.000 & 1.000 & 0.010 & 0.003 & \rbrk{$1.0\%$}  & 0.972 & \bad \\
      & \textbf{\method{}} ($\delta{=}4$)   & 0.996 & 0.975 & 0.943 & 0.850 & \rsurv{$94.7\%$} & --- & \good \\
      & \textbf{\method{}} ($\delta{=}8$)   & 0.991 & 0.968 & 0.852 & 0.698 & \rsurv{$86.0\%$} & --- & \good \\
      & CM ($\delta{=}5, k{=}64$)           & 1.000 & 1.000 & 1.000 & 0.997 & \rsurv{$100\%$}  & --- & \good \\
      & CM ($\delta{=}8, k{=}64$)           & 1.000 & 1.000 & 1.000 & 1.000 & \rsurv{$100\%$}  & --- & \good \\
      & IndexMark ($\tau{=}0$)              & 1.000 & 1.000 & 0.000 & 0.000 & \rbrk{$0.0\%$}  & 0.948 & \bad \\
      & IndexMark ($\tau{=}0.8$)            & 0.928 & 0.420 & 0.060 & 0.000 & \rbrk{$6.5\%$}  & 0.480 & \bad \\
\midrule
\multirow{8}{*}{LlamaGen-L}
      & KGW $h{=}0$ ($\delta{=}4$)          & 1.000 & 1.000 & 0.007 & 0.000 & \rbrk{$0.7\%$}  & 0.959 & \bad \\
      & KGW $h{=}0$ ($\delta{=}8$)          & 0.995 & 0.990 & 0.024 & 0.007 & \rbrk{$2.4\%$}  & 0.959 & \bad \\
      & \textbf{\method{}} ($\delta{=}4$)   & 1.000 & 1.000 & 0.965 & 0.773 & \rsurv{$96.5\%$} & --- & \good \\
      & \textbf{\method{}} ($\delta{=}8$)   & 0.995 & 0.990 & 0.935 & 0.557 & \rsurv{$94.0\%$} & --- & \good \\
      & CM ($\delta{=}5, k{=}64$)           & 1.000 & 1.000 & 0.995 & 0.970 & \rsurv{$99.5\%$} & --- & \good \\
      & CM ($\delta{=}8, k{=}64$)           & 0.999 & 1.000 & 0.996 & 0.993 & \rsurv{$99.7\%$} & --- & \good \\
      & IndexMark ($\tau{=}0$)              & 1.000 & 1.000 & 0.005 & 0.003 & \rbrk{$0.5\%$}  & 0.950 & \bad \\
      & IndexMark ($\tau{=}0.8$)            & 0.911 & 0.367 & 0.073 & 0.000 & \rbrk{$8.0\%$}  & 0.465 & \bad \\
\bottomrule
\end{tabular}
\end{table*}

\subsection{Why frequency counting recovers a static greenlist}
\label{supp:freq}

\begin{figure}[!htb]
\centering
\includegraphics[width=\columnwidth]{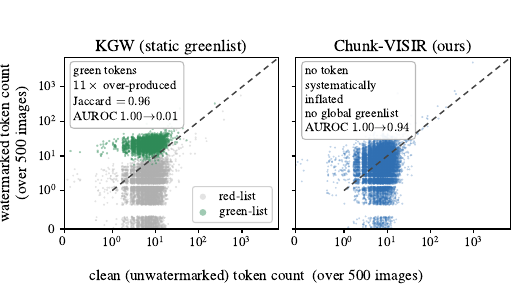}
\caption{\textbf{Why frequency-counting recovers a static greenlist but
not \method{}'s.} Watermarked vs.\ clean per-token count over
$N_\mathrm{atk}{=}500$ LlamaGen-B images ($\delta{=}4$); each dot is a
codebook entry, and the dashed line is $y{=}x$. \textbf{KGW $h{=}0$:}
greenlist tokens (\textcolor{gainGreen}{green}) detach above the
diagonal (up to ${\sim}11\times$ their clean count), so a top-$\lceil\gamma V\rceil$
ranking recovers the greenlist (Jaccard $0.96$) and removal drops AUROC to
$0.015$, matching the ${\sim}3\times$ rank-averaged cliff of Fig.~3c.
\textbf{\method:} the partition moves with the trajectory, counts stay on
the diagonal, and post-removal AUROC remains $0.943$.}
\label{fig:removal_freq}
\end{figure}

\cref{tab:removal_full} reports what the removal attack does to each
detector, while \cref{fig:removal_freq} shows why directly at the raw
token counts observed by the attacker without a detector in the loop. A
scheme with one global greenlist over-produces exactly the tokens on it,
so those entries detach above the diagonal and a simple count ranking
recovers the partition. When the partition varies along the generation
trajectory under $\lpc$, marginal token counts remain balanced along the
diagonal, leaving frequency ranking uninformative (Tab.~1).

\subsection{Attacker-budget sensitivity}
\label{supp:natk}

The adaptive removal attack of~Sec.~4.3 uses a canonical budget
of $N_\mathrm{atk}{=}500$ images. To test whether larger query budgets
erode secrecy, we sweep $N_\mathrm{atk}\in\{100, 300, 500, 1000, 2000,
5000\}$, spanning a $50\times$ range up to $10\times$ the main-paper
budget (\cref{tab:natk_full}). Three clear patterns emerge: KGW $h{=}0$
collapses already at $N_\mathrm{atk}{=}100$ ($0.066$ AUROC) and reaches
$0.986$ Jaccard overlap by $2\,000$ queries ($0.009$ AUROC at $5\,000$);
\method{} remains flat across the entire $50\times$ span ($0.949\pm 0.020$
mean AUROC, retaining $0.924$ even at $5\,000$ queries, with minor
cell-to-cell differences reflecting finite-sample evaluation variance);
and ClusterMark likewise holds at $0.995$ to $1.000$. Thus, increasing the
attacker's query budget by an order of magnitude produces no asymptotic
erosion of \method{}'s removal resistance.

We also vary both query budget and cluster count in the context-aware
rule-inference attack (\cref{tab:ri_sweep}). Across a $20\times$ range of
query budgets and a $16\times$ range of cluster counts, balanced accuracy
stays tightly bounded near chance ($0.52$ to $0.58$ at full or near-full
coverage; at $N_\mathrm{atk}{=}100$ with $256$ clusters, sparse bucket
counts cover only $16.6\%$ of contexts), confirming that rule-inference
resistance holds across quantization granularities.

\begin{table}[!htb]
\centering
\footnotesize
\setlength{\tabcolsep}{3.5pt}
\caption{Rule-inference balanced accuracy against \method{} (chance
$0.5$) across attacker query budgets and context cluster counts, with
estimable context coverage in parentheses (evaluated on a shared generated
split; $0.540$ at $N_\mathrm{atk}{=}500, k{=}64$ matches the $0.536$
headline in Tab.~1 within sample variation).}
\label{tab:ri_sweep}
\begin{tabular}{@{}lccc@{}}
\toprule
$N_\mathrm{atk}$ & $16$ clusters & $64$ clusters & $256$ clusters \\
\midrule
$100$  & 0.535 \tiny(1.000) & 0.524 \tiny(0.939) & 0.626 \tiny(0.166) \\
$500$  & 0.562 \tiny(1.000) & \textbf{0.540} \tiny(1.000) & 0.537 \tiny(0.980) \\
$2000$ & 0.581 \tiny(1.000) & 0.558 \tiny(1.000) & 0.544 \tiny(1.000) \\
\bottomrule
\end{tabular}
\end{table}

\begin{table}[!htb]
\centering
\scriptsize
\setlength{\tabcolsep}{2.8pt}
\caption{Removal AUROC on LlamaGen-B ($\delta{=}4$) across a $50\times$
attacker query budget $N_\mathrm{atk}$. ``Jacc.'' denotes inferred-vs-true
greenlist Jaccard for KGW $h{=}0$. \method{} holds flat at $0.949\pm 0.020$
with no downward trend through $5\,000$ queries. The sweep is an
independent run family evaluated on its own generated splits, so its
$N_\mathrm{atk}{=}500$ cells differ from the canonical run of
\cref{tab:removal_full} ($0.943$ AUROC, $0.959$ Jaccard) by run-to-run
variation only.}
\label{tab:natk_full}
\begin{tabular}{@{}lcccccc@{}}
\toprule
$N_\mathrm{atk}$ & 100 & 300 & 500 & 1000 & 2000 & 5000 \\
\midrule
KGW $h{=}0$ AUROC & 0.066 & 0.016 & 0.016 & 0.009 & 0.005 & 0.009 \\
KGW $h{=}0$ Jacc. & 0.804 & 0.957 & 0.982 & 0.984 & 0.986 & 0.985 \\
\method{} AUROC   & 0.931 & 0.978 & 0.941 & 0.965 & 0.954 & 0.924 \\
ClusterMark AUROC & 0.998 & 0.999 & 1.000 & 0.999 & 0.995 & 0.996 \\
\bottomrule
\end{tabular}
\end{table}

\subsection{Cluster-aware adaptive attack on ClusterMark}
\label{sec:supp_cm_aware}

The global frequency-removal attack (\cref{tab:removal_full}) uses a single
cluster-oblivious frequency table. To stress-test ClusterMark more
severely, we give the attacker its own $k$-means pass on the VQ codebook
($k{=}64$), align it to ClusterMark's partition up to a label permutation,
estimate a per-cluster greenlist from $500$ queries, and replace each
green token with its nearest-neighbor non-green token \emph{within the
same cluster}. \cref{tab:cm_aware_supp} shows that even when attacker and
defender partitions coincide exactly (Rand index~$=1.00$), post-attack
AUROC drops by at most $0.9$\,pp. Splitting the codebook into $64$
clusters raises the query cost of within-cluster replacement roughly in
proportion, explaining ClusterMark's removal survival. However, because
its rule depends only on a single preceding token's discrete cluster ID
over a public codebook, ClusterMark trails \method{} on $10$ of the $16$
geometric and rescale cells in \cref{tab:geom_full} (tying on $4$) due to
cross-cluster token drift, and its recoverable $64$-cluster partition
leaves it directly vulnerable to watermark forgery (\cref{supp:forgery}).

\begin{table}[!htb]
\centering
\footnotesize
\setlength{\tabcolsep}{3.5pt}
\caption{Cluster-aware adaptive removal attack on ClusterMark. \emph{Rand}
is the pair-agreement Rand index between the attacker's and defender's
partitions. The attack replaces $33\%$ to $44\%$ of tokens, yet AUROC
drops by at most $0.9$\,pp.}
\label{tab:cm_aware_supp}
\begin{tabular}{@{}llccccc@{}}
\toprule
Model & Seed & Rand & Replaced & Clean & After & Drop \\
\midrule
B & aligned & 1.000 & 43.7\% & 1.000 & 0.996 & 0.004 \\
B & off-A   & 0.087 & 33.0\% & 1.000 & 0.999 & 0.001 \\
B & off-B   & 0.085 & 33.3\% & 1.000 & 0.999 & 0.001 \\
L & aligned & 1.000 & 43.7\% & 1.000 & 0.991 & 0.009 \\
L & off-A   & 0.087 & 33.4\% & 1.000 & 0.998 & 0.002 \\
L & off-B   & 0.085 & 33.4\% & 1.000 & 0.997 & 0.003 \\
\bottomrule
\end{tabular}
\end{table}

\subsection{Forgery: what a recovered rule is worth}
\label{supp:forgery}

While \cref{sec:supp_cm_aware} shows that partitioning the codebook into
$64$ clusters slows down naive token-replacement removal, rule secrecy in
a provenance system must also defend against watermark \emph{forgery}
(spoofing). A rule that can be inferred from queried outputs can be
replayed onto arbitrary unwatermarked images, causing the detector to
falsely attribute external or malicious content to the protected
generator. For provenance verification, false attribution is just as
damaging as watermark removal, and it directly tests whether the
conditioning rule itself remains secret.

We evaluate left-to-right forgery against both ClusterMark and \method{}
under a matched $500$-query budget and estimator family. From $500$
watermarked images, the attacker estimates a per-bucket greenlist via
frequency counting and rewrites $300$ \emph{unwatermarked} images left to
right, replacing each token with the nearest codebook entry inside the
inferred greenlist selected by its already-forged predecessor. The forged
tokens are decoded back to pixels and scored by each scheme's own detector
after the VQ round trip ($78\%$ of forged tokens survive re-encoding for
both schemes). ClusterMark's greenlist is a union of $k$-means codebook
clusters, collapsing each bucket's rule from $V{=}16\,384$ binary
decisions to $64$ cluster-level choices; \method{}'s greenlist is an
arbitrary vocabulary subset indexed by a continuous trajectory summary and
admits no such low-dimensional compression.

\cref{tab:forgery} shows a decisive contrast. Against ClusterMark,
$96.3\%$ of forged images pass the detector's $1\%$-FPR threshold, with a
mean score ($+16.8$) close to genuine watermarked images ($+19.9$), making
forgeries virtually indistinguishable from authentic outputs. Against
\method{}, the identical attack achieves only a $2.0\%$ pass rate (mean
score $-0.27$, near clean negatives at $-0.31$), effectively preventing
spoofing.

This result underscores the operational importance of rule-inference
resistance (Tab.~1): static codebook clustering resists
removal yet exposes a compact $64$-cluster rule that enables $96.3\%$
forgery, whereas \method{}'s continuous trajectory conditioning blocks
both removal and spoofing.

\begin{table}[!htb]
\centering
\footnotesize
\setlength{\tabcolsep}{4pt}
\caption{Forgery attack on LlamaGen-B. The attacker queries $500$
watermarked images and applies the estimated rule to $300$ unwatermarked
images. \emph{Forged} rows report attacker pass rates and detector scores
after the VQ round trip; \emph{Reference} rows verify genuine and negative
baselines on the same detector and threshold.}
\label{tab:forgery}
\begin{tabular}{@{}lcc@{}}
\toprule
 & ClusterMark & \oc{\method{}} \\
\midrule
\multicolumn{3}{@{}l}{\itshape Forged images (attacker succeeds if high)} \\
\quad TPR at $1\%$ FPR            & \rbrk{0.963} & \oc{\rsurv{\best{0.020}}} \\
\quad mean detector score         & \rbrk{$+16.8$} & \oc{$-0.27$} \\
\midrule
\multicolumn{3}{@{}l}{\itshape Reference (same detector and threshold)} \\
\quad genuine TPR at $1\%$ FPR    & 1.000 & \oc{0.970} \\
\quad mean score, genuine         & $+19.9$ & \oc{$+0.27$} \\
\quad mean score, negatives       & $-0.37$ & \oc{$-0.31$} \\
\bottomrule
\end{tabular}
\end{table}

\subsection{A continuous-surrogate attacker}
\label{supp:surrogate}

Frequency counting estimates one global greenlist (Sec.~4.3.1),
and clustering-based rule inference quantizes the conditioning space into
discrete buckets (Sec.~4.3.2). Because \method{}'s detector
averages a continuous score over emitted tokens, one might ask whether a
neural attacker could instead fit a smooth function that correlates with
$T_\theta$'s continuous bias directly on the unquantized conditioning
space and drive detector scores below threshold without ever recovering a
discrete partition. We test this continuous-surrogate threat model
directly.

\paragraph{Attacker formulation.}
Under the threat model of Sec.~3, the attacker knows the
architecture and public VQ codebook (but not $\theta$ or $\textit{key}$),
and can recompute $\mathbf{c}_{k-1}$ from any observed token sequence
after the key-seeded first chunk. From $500$ queried watermarked images,
the attacker trains a continuous surrogate
$w_\phi(\mathbf{c}_{k-1})\!\in\!\mathbb{R}^V$ by ranking each emitted
token above a negative token sampled from the unwatermarked base
distribution. At attack time, it recomputes $\mathbf{c}_{k-1}$ on the
target image and replaces up to a budget fraction of tokens with whichever
of their $16$ nearest codebook neighbors minimizes $w_\phi$. Because the
canonical frequency-counting attack of \cref{tab:removal_full} rewrites
$50.0\%$ of emitted positions in practice, a $0.50$ rewrite cap provides
an exact budget-matched comparison alongside LPIPS and PSNR distortion.

\paragraph{Results.}
The base continuous surrogate reaches $0.535$ greenlist balanced accuracy
(matching $0.536$ under $k$-means clustering), yet converts this into
\emph{less} watermark removal than discrete frequency counting
(\cref{tab:surrogate}). At the matched $0.50$ token budget, both attacks
incur nearly identical visual distortion (LPIPS $0.080$ vs.\ $0.068$, PSNR
$24.9$ vs.\ $25.0$\,dB), where frequency counting reaches $0.943$ AUROC
while the continuous surrogate leaves detection at $0.984$ AUROC ($0.963$
$\TPR@1\%$). Examining the raw detector scores explains why: at budget
$0.25$, the mean watermarked score drops from $+0.26$ to $+0.06$, but the
mean clean negative score simultaneously shifts from $-0.31$ to $-0.35$.
Rather than isolating the secret key-permuted partition, $w_\phi$ captures
shared generator token preferences and shifts both distributions downward
in parallel without collapsing their separation. Across all
image-preserving budgets ($0.10$ to $0.50$, $\text{LPIPS}\le 0.080$),
post-attack AUROC stays between $0.984$ and $0.990$. Pushing the rewrite
budget to $1.00$ ($96.8\%$ of all tokens replaced) still leaves detection
well above chance ($0.778$ AUROC vs.\ $0.015$ for frequency counting on
KGW $h{=}0$) only by inflicting severe perceptual degradation (LPIPS
$0.146$, PSNR $22.2$\,dB) while rule recovery remains at $0.535$.

\begin{table*}[!t]
\centering
\footnotesize
\setlength{\tabcolsep}{3.8pt}
\caption{Continuous-surrogate removal against \method{} on LlamaGen-B
($500$ attacker queries, $300$ test images, $\delta{=}4$). Budget is the
fraction of positions the attacker may rewrite ($0.50$ matches the
$50.0\%$ empirical rewrite rate of frequency counting in
\cref{tab:removal_full}). LPIPS and PSNR are measured against the
unattacked watermarked image. The continuous surrogate removes less
watermark than plain frequency counting at matched perceptual distortion;
KGW $h{=}0$ under frequency counting is shown for reference.}
\label{tab:surrogate}
\begin{tabular}{@{}lccccccc@{}}
\toprule
Attack & \makecell{token\\budget} & \makecell{tokens\\changed} & AUROC $\uparrow$
       & $\TPR@1\%$ $\uparrow$ & \makecell{mean score\\wm\,/\,clean}
       & LPIPS $\downarrow$ & PSNR $\uparrow$ \\
\midrule
none (clean)                 & ---    & ---      & 1.000$^\dagger$ & 0.987 & $+0.26$\,/\,$-0.31$ & ---   & ---  \\
\midrule
frequency counting           & ---    & $50.0\%$ & 0.943$^\ddagger$ & 0.850 & ---  & 0.068 & 25.0 \\
\quad \textit{vs.\ KGW $h{=}0$} & --- & $82.0\%$ & \textcolor{red}{0.015}$^\ddagger$ & 0.007 & --- & 0.123 & 21.6 \\
\midrule
surrogate ($\mathbf{c}_{k-1}$) & $0.10$ & $10.2\%$ & 0.990 & 0.987 & $+0.17$\,/\,$-0.33$ & 0.021 & 31.4 \\
surrogate ($\mathbf{c}_{k-1}$) & $0.25$ & $25.0\%$ & 0.988 & 0.980 & $+0.06$\,/\,$-0.35$ & 0.045 & 27.6 \\
surrogate ($\mathbf{c}_{k-1}$) & $0.50$ & $50.0\%$ & \textbf{0.984} & 0.963 & $-0.11$\,/\,$-0.38$ & 0.080 & 24.9 \\
surrogate ($\mathbf{c}_{k-1}$) & $1.00$ & $96.8\%$ & 0.778 & 0.083 & $-0.34$\,/\,$-0.39$ & 0.146 & 22.2 \\
\bottomrule
\end{tabular}
\\[2pt]
{\footnotesize $^\dagger$ $0.991$ on this $300$-image test split.
$^\ddagger$ canonical LlamaGen-B run of \cref{tab:removal_full}; perceptual
metrics are measured on a matched re-run ($0.966$ AUROC on \method{},
$0.014$ on KGW), while retaining the conservative $0.943$ figure
throughout the paper.}
\end{table*}

\subsection{Multi-key secrecy}
\label{sec:supp_multikey}

A deployed provenance system must support multiple independent keys so
that compromising one key does not expose others. As stated in
Eq.~(4) and \cref{alg:chunkvisir}, \method{} derives an
arbitrary number of keys from a single trained $T_\theta$ without
retraining by indexing $T_\theta$'s $V$-dimensional output vector at
\emph{every} position $t\in\{1,\dots,N\}$ through a key-seeded vocabulary
permutation $\sigma_{\textit{key}}\in S_V$:
$w_t^{(\textit{key})}(v) = w_t^{\text{base}}(\sigma_{\textit{key}}(v))$.
Both generation and verification apply $\sigma_{\textit{key}}$ across all
chunks, so replaying verification under a mismatched key $\textit{key}'\neq
\textit{key}$ permutes the vocabulary alignment at every token position
rather than only in the first chunk.

\cref{tab:multikey} evaluates this vocabulary-permutation mechanism across
four independently seeded $64$-bit keys ($240$ test images per key,
$\delta{=}8$) on LlamaGen-B, calibrating each detector key's $1\%$-FPR
threshold on $240$ clean unwatermarked images ($99$th empirical quantile).
On-diagonal detection reaches $1.000$ TPR with a mean detector score of
$+0.392\pm 0.140$. Off-diagonal (mismatched-key) scores average
$-0.314\pm 0.036$, matching the unwatermarked clean baseline distribution
($-0.319\pm 0.033$) within $0.15$ standard deviations and separated from
genuine on-diagonal scores by nearly $20$ negative standard deviations
(with empirical off-diagonal pass rates averaging $0.025$ due to
finite-sample quantile estimation on $240$ clean images). Output
vocabulary permutation thus isolates keys across the entire token sequence
at zero retraining cost.

\begin{table}[!htb]
\centering
\footnotesize
\caption{Cross-key detection TPR at $\FPR{=}1\%$ and mean detector scores
with vocabulary permutation $\sigma_{\textit{key}}$ enabled at every
position across four keys on LlamaGen-B ($240$ images per key,
$\delta{=}8$). Diagonal TPR is $1.000$ (mean score $+0.392$), whereas
mismatched-key scores ($-0.314$) match clean unwatermarked images
($-0.319$).}
\label{tab:multikey}
\setlength{\tabcolsep}{3.0pt}
\begin{tabular}{@{}lccccc@{}}
\toprule
gen $\backslash$ det & 2024 & 1337 & 42 & 7 & clean \\
\midrule
2024  & \best{1.000} & 0.038        & 0.021        & 0.025        & --- \\
1337  & 0.017        & \best{1.000} & 0.008        & 0.021        & --- \\
42    & 0.029        & 0.050        & \best{1.000} & 0.013        & --- \\
7     & 0.029        & 0.038        & 0.013        & \best{1.000} & --- \\
clean & 0.013        & 0.013        & 0.013        & 0.013        & --- \\
\midrule
on-diag mean   & $+0.398$ & $+0.389$ & $+0.401$ & $+0.378$ & --- \\
off-diag mean  & $-0.314$ & $-0.307$ & $-0.318$ & $-0.316$ & --- \\
clean mean     & $-0.324$ & $-0.315$ & $-0.319$ & $-0.317$ & --- \\
\bottomrule
\end{tabular}
\end{table}

\paragraph{Non-default keys at canonical $\delta{=}4$: isolation, post-edit robustness, and removal.}
To verify that applying non-identity permutations $\sigma_{\textit{key}}\in S_V$
in Eq.~(4) to a single trained $T_\theta$ preserves all canonical
$\delta{=}4$ properties without retraining, \cref{tab:multikey_d4}
evaluates four non-default keys alongside the canonical base key ($2024$)
on LlamaGen-B ($240$ images per key, $N_\mathrm{atk}{=}500$): three
independent uniform pseudorandom permutations ($1337$, $42$, $7$) and a
geometry-preserving isometric codebook permutation (\emph{1337-iso},
induced by a key-seeded orthogonal rotation in codebook space). Across the
four standard keys, on-diagonal clean detection reaches $0.999$ to $1.000$
AUROC ($0.992$ to $1.000$ $\TPR@1\%$), whereas all twelve off-diagonal key
pairs stay near chance ($0.532$ mean AUROC; mean scores $-0.308$ to
$-0.323$, matching clean negatives at $-0.316$ to $-0.321$). Across all
eight aligned edits of Tab.~1, every non-default key
reaches $0.988$ to $0.992$ mean AUROC ($0.948$ to $0.953$ $\TPR@1\%$),
matching the canonical base key because input-conditioning smoothness
($\mathcal{L}_s$) is invariant to output permutation. Under misaligning
$5\%$ cropping, uniform permutations retain $0.918$ to $0.927$ AUROC
(well above KGW $h{=}1$ at $0.804$), while \emph{1337-iso} additionally
preserves codeword-neighborhood smoothness ($\mathcal{L}_\mathrm{cb}$) and
reaches $0.949$ AUROC ($0.642$ $\TPR@1\%$). Under the $500$-query
frequency-removal attack, all four non-default keys retain $0.987$ to
$0.998$ AUROC ($0.942$ to $0.971$ $\TPR@1\%$). Even under a public-trunk
threat model where all weights of $g_\theta$ are known and only
$\sigma_{\textit{key}}$ is secret, $500$ watermarked queries recover only
$0.23\%$ to $0.28\%$ of the $V{=}16\,384$ permutation entries.

\begin{table}[!htb]
\centering
\scriptsize
\setlength{\tabcolsep}{2.0pt}
\caption{Non-default key verification on LlamaGen-B at the canonical
$\delta{=}4$ operating point ($240$ images per key, $N_\mathrm{atk}{=}500$)
without retraining $T_\theta$. Cells report AUROC/matched $\TPR@1\%$;
\emph{Leak} is exact $\sigma_{\textit{key}}$ recovery when $g_\theta$ is public.}
\label{tab:multikey_d4}
\begin{tabular}{@{}lccccc@{}}
\toprule
Key & Clean & Aligned & Crop $5\%$ & Removed & Leak \\
\midrule
$2024$\,(ref.)$^\dagger$ & 1.000/1.000 & 0.993/0.965 & 0.961/0.682 & 0.943/0.850 & --- \\
$1337$                   & 1.000/1.000 & 0.991/0.949 & 0.922/0.450 & 0.998/0.950 & 0.26\% \\
$42$                     & 1.000/1.000 & 0.992/0.953 & 0.927/0.642 & 0.997/0.963 & 0.28\% \\
$7$                      & 0.999/0.992 & 0.988/0.948 & 0.918/0.525 & 0.997/0.971 & 0.25\% \\
$1337$-iso               & 1.000/1.000 & 0.992/0.948 & 0.949/0.642 & 0.987/0.942 & 0.23\% \\
\bottomrule
\end{tabular}
\\[2pt]
{\footnotesize $^\dagger$Canonical $400$-image benchmark (Tab.~1 and \cref{tab:removal_full}); on this $240$-image split, key $2024$ reads $0.999/0.996$ clean, $0.972/0.907$ aligned, and $0.949/0.463$ crop.}
\end{table}

\section{Ablations, Transferability, Fidelity, and Runtime}
\label{sec:supp_abl_sys}

\subsection{Ablation analysis}
\label{sec:supp_ablation}

We ablate eight design choices by training a fresh $T_\theta$ for each
variant on LlamaGen-B with all other hyperparameters held fixed: chunk
size ($C\in\{16, 64\}$ vs.\ $32$), removing the local preceding-token
branch (\emph{w/o local branch}), removing individual loss terms (\lpc,
\lgf, or \lent), output sharpness $k_2{=}10$, and clean-only training
without differentiable attack augmentation (\cref{fig:design}b,
\cref{tab:ablation_removal}).

Removing \lpc{} dominates the ablation results, moving \method{} along the
same trade-off axis that separates the baselines in
Tab.~1. Dropping \lpc{} reduces post-removal AUROC to
$0.285$ ($0.965 \to 0.313$ under the three-seed matched ablation protocol
of \cref{tab:lpc_pareto}) while raising average image-attack AUROC to
$0.949$ over this run's evaluation grid ($0.957 \pm 0.001$ on the full
twenty-one-attack suite under the same matched protocol), making it the
only variant vulnerable to frequency-based removal
as positions collapse toward a static global greenlist akin to $h{=}0$.
Thus, \lpc{} directly governs the robustness-secrecy trade-off of
Sec.~3.2 within a single architecture; we adopt
$\lambda_\mathrm{pc}{=}2.5$ because it clears the removal-resistance
ceiling while retaining $0.993$ aligned-edit AUROC (\cref{supp:pareto}).

Across all remaining seven variants, post-removal AUROC stays consistently
high ($\geq 0.916$). On the image-transformation side, three components
provide clear gains in our default dual-stream configuration: (i) adding
the local token branch on top of the pooled $C{=}32$ chunk summary boosts
geometric sensitivity ($5^\circ$ rotation AUROC rises from $0.796$ to
$0.822$, and horizontal flip from $0.677$ to $0.742$) while keeping
clustering rule-inference accuracy near chance ($0.540$ vs.\ $0.542$ at
$64$ clusters); (ii) shrinking the chunk to $C{=}16$ degrades average
image-attack AUROC by $0.028$ (including $0.614$ vs.\ $0.742$ on
horizontal flip), confirming the benefit of pooling over $C{=}32$ tokens;
and (iii) removing differentiable attack augmentation lowers average
image-attack AUROC by $0.050$, validating our training pipeline.

\begin{table}[!htb]
\centering
\footnotesize
\caption{Eight-way ablation on the adaptive removal attack on LlamaGen-B
(companion to \cref{fig:design}b). Removing \lpc{} is the only change
that breaks removal resistance; all other variants remain above $0.916$
AUROC (within normal run-to-run and seed variance; see \cref{tab:natk_full,tab:lpc_pareto}),
while \lgf{} and \lent{} regulate signal scale and saturation
(\cref{supp:implementation,sec:supp_gamma}) and the full model achieves
the highest $5^\circ$-rotation AUROC among all removal-resistant variants.}
\label{tab:ablation_removal}
\setlength{\tabcolsep}{4.0pt}
\begin{tabular}{@{}lcccc@{}}
\toprule
Variant                        & clean & \textbf{removed} & rot $5^\circ$ & verdict \\
\midrule
no \lpc                        & 0.981 & \textcolor{red}{0.285} & 0.965 & broken \\
no \lgf                        & 0.993 & 0.916 & 0.820 & OK \\
main (full, $C{=}32$)          & 0.996 & 0.943 & \textbf{0.822} & \textbf{adopted} \\
chunk 64                       & 1.000 & 0.963 & 0.793 & OK \\
$k_2=10$                       & 1.000 & 0.972 & 0.807 & OK \\
w/o local branch               & 0.999 & 0.979 & 0.796 & OK \\
no attack-aug                  & 1.000 & 0.982 & 0.786 & OK \\
chunk 16                       & 0.999 & 0.984 & 0.786 & OK \\
no \lent                       & 1.000 & 1.000 & 0.819 & OK \\
\bottomrule
\end{tabular}
\end{table}

\paragraph{Decomposing the smoothness regularizer ($\mathcal{L}_s$ vs.\ $\mathcal{L}_\mathrm{cb}$).}
In Eq.~(5), smoothness under VQ drift (R1) is jointly encouraged by
end-to-end differentiable attack augmentation through soft VQ re-encoding
($\ldet$) and the explicit regularizer
$\mathcal{L}_\mathrm{sm}=\mathcal{L}_s+\mathcal{L}_\mathrm{cb}$. To
isolate the complementary roles of input-cosine alignment ($\mathcal{L}_s$)
and codeword-neighborhood smoothness ($\mathcal{L}_\mathrm{cb}$),
\cref{tab:smooth_decomp} compares the three-seed full model from the
matched ablation recipe of \cref{tab:lpc_pareto} ($\lambda_\mathrm{pc}{=}2.5$)
against retraining without $\mathcal{L}_s$ (\emph{w/o $\mathcal{L}_s$}) or
without $\mathcal{L}_\mathrm{cb}$ (\emph{w/o $\mathcal{L}_\mathrm{cb}$}). On
in-distribution $256^2$ global AUROC under the base key, differentiable
attack augmentation largely masks single-term removal ($5^\circ$-rotation
AUROC reads $0.837\pm 0.003$ vs.\ $0.846$ without $\mathcal{L}_s$ and
$0.824$ without $\mathcal{L}_\mathrm{cb}$, with post-removal AUROC at
$0.965$ to $0.969$). Under strict low-FPR thresholds and out-of-distribution
deployments, however, each term prevents a distinct failure mode:
(i) dropping $\mathcal{L}_\mathrm{cb}$ raises $8$-NN codeword bias MSE by
$55\%$ ($0.323\to 0.500$) and sharply degrades $\TPR@1\%\FPR$ whenever
edits shift tokens to neighboring codewords, losing $-21.9$\,pp on $5\%$
center cropping ($0.756\to 0.537$), $-23.8$\,pp on $25\%$ downscaling
($0.718\to 0.480$), $-22.6$\,pp on Gaussian noise $\sigma{=}0.10$
($0.816\to 0.590$), and $-9.9$\,pp on blur $r{=}4$ ($0.904\to 0.805$),
while shrinking active vocabulary coverage ($14\,922\to 14\,581$ codes);
(ii) dropping $\mathcal{L}_s$ degrades generalization whenever output
permutations or resolution shifts alter trajectory statistics: under a
non-default secret key ($\textit{key}{=}1337$), $5^\circ$-rotation AUROC
drops by $-6.1$\,pp ($0.754\to 0.693$) and JPEG $q25$ AUROC falls from
$0.995$ to $0.972$; under zero-shot $384{\times}384$ transfer ($576$
tokens), aligned-edit mean AUROC ($0.942\to 0.926$), $5\%$ cropping
($0.876\to 0.843$), and $5^\circ$ rotation ($0.823\to 0.800$) all drop
below every full-model seed; and even at $256^2$ under the base key,
$5\%$-crop $\TPR@1\%$ drops by $-8.4$\,pp ($0.756\to 0.672$). Combining
$\mathcal{L}_s$ and $\mathcal{L}_\mathrm{cb}$ is therefore the only
configuration that avoids severe drops across low-FPR recall, non-default
keys, and zero-shot cross-resolution transfer simultaneously.

\begin{table*}[!t]
\centering
\footnotesize
\setlength{\tabcolsep}{3.2pt}
\caption{Multi-dimensional decomposition of the smoothness objective
$\mathcal{L}_\mathrm{sm}=\mathcal{L}_s+\mathcal{L}_\mathrm{cb}$ on
LlamaGen-B ($\delta{=}4$, matched ablation recipe; full model reports mean
$\pm$ std across the three $\lambda_\mathrm{pc}{=}2.5$ seeds of
\cref{tab:lpc_pareto}). \textcolor{lossRed}{Red} marks substantial
degradations relative to the full model ($>1.5$ seed std);
Full ($\mathcal{L}_s{+}\mathcal{L}_\mathrm{cb}$) is the only configuration
with no failure mode across strict low-FPR recall, non-default keys, and
zero-shot $384^2$ transfer.}
\label{tab:smooth_decomp}
\begin{tabular}{@{}l ccc cc cc@{}}
\toprule
 & \multicolumn{3}{c}{$256^2$ low-FPR recall ($\TPR@1\%$)}
 & \multicolumn{2}{c}{Non-default $\textit{key}{=}1337$ AUROC}
 & \multicolumn{2}{c}{Zero-shot $384^2$ AUROC} \\
\cmidrule(lr){2-4}\cmidrule(lr){5-6}\cmidrule(lr){7-8}
Variant
 & Crop $5\%$ & Resize $25\%$ & Noise $0.10$
 & JPEG $q25$ & Rot $5^\circ$
 & Aligned (8) & Crop $5\%$ \\
\midrule
\textbf{Full ($\mathcal{L}_s{+}\mathcal{L}_\mathrm{cb}$)}
 & $0.756{\pm}.047$ & $0.718{\pm}.050$ & $0.816{\pm}.056$
 & $0.995{\pm}.004$ & $0.754{\pm}.038$
 & $0.942{\pm}.008$ & $0.876{\pm}.017$ \\
w/o $\mathcal{L}_s$
 & \textcolor{lossRed}{0.672} & 0.725                      & 0.828
 & \textcolor{lossRed}{0.972} & \textcolor{lossRed}{0.693}
 & \textcolor{lossRed}{0.926} & \textcolor{lossRed}{0.843} \\
w/o $\mathcal{L}_\mathrm{cb}$
 & \textcolor{lossRed}{0.537} & \textcolor{lossRed}{0.480} & \textcolor{lossRed}{0.590}
 & 0.989                      & 0.719
 & 0.968                      & 0.878 \\
\bottomrule
\end{tabular}
\end{table*}

\subsection{Green-fraction ($\gamma$) sweep}
\label{sec:supp_gamma}

$\gamma$ sets the target vocabulary fraction promoted at each step.
Because \lgf{} is a one-sided lower-bound barrier holding the positive
fraction at or above its natural $0.25$ equilibrium, \cref{tab:gamma}
sweeps $\gamma\in\{0.25, 0.40, 0.50\}$ under a shared training schedule
with all other hyperparameters fixed. Clean detection, post-removal AUROC,
and average image-attack AUROC all peak at $\gamma{=}0.25$ and decline as
denser greenlists dilute per-token selectivity, validating our default
choice.

\begin{table}[!htb]
\centering
\footnotesize
\setlength{\tabcolsep}{5pt}
\caption{Green-fraction sweep on LlamaGen-B ($\delta{=}4$, shared sweep
schedule). Because \lgf{} is a one-sided lower-bound barrier, $\gamma$ is
swept upward from $0.25$; denser greenlists weaken clean detection,
removal resistance, and image-attack robustness alike.}
\label{tab:gamma}
\begin{tabular}{@{}lccc@{}}
\toprule
$\gamma$ & clean & \textbf{removed} & image-attack avg \\
\midrule
$0.25$ (default) & \best{0.995} & \best{0.989} & \best{0.884} \\
$0.40$           & 0.991 & 0.961 & 0.870 \\
$0.50$           & 0.979 & 0.926 & 0.872 \\
\bottomrule
\end{tabular}
\end{table}

\subsection{Pareto curve for the removal/geometric trade-off}
\label{supp:pareto}

\begin{figure*}[!t]
\centering
\includegraphics[width=\textwidth]{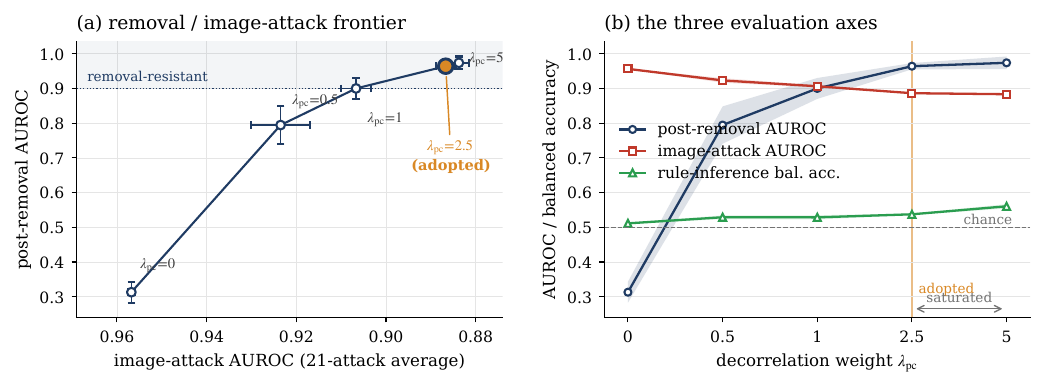}
\caption{\textbf{The decorrelation weight $\lambda_\mathrm{pc}$ traces a
monotone Pareto frontier.} LlamaGen-B, $\delta{=}4$, three seeds per
setting; error bars denote one standard deviation.
\textbf{(a)}~Post-removal AUROC against average AUROC over the
twenty-one-attack image-level suite (the $x$~axis is reversed so that
both axes improve upward and rightward); $\lambda_\mathrm{pc}{=}2.5$ sits
at the knee where removal resistance saturates.
\textbf{(b)}~The same runs across all three evaluation axes, showing how
post-removal AUROC and spatial robustness determine the operating point
while rule-inference accuracy remains near chance everywhere.}
\label{fig:lpc_pareto}
\end{figure*}

\cref{tab:lpc_pareto} and \cref{fig:lpc_pareto} report a dedicated sweep
over $\lambda_\mathrm{pc}\in\{0, 0.5, 1.0, 2.5, 5.0\}$ on LlamaGen-B
($\delta{=}4$), holding all other training hyperparameters fixed and
retraining each setting across three independent seeds under the shared
ablation schedule of \cref{sec:supp_ablation}. Each run is evaluated on
three axes: post-removal AUROC (operational secrecy), average AUROC over
the twenty-one-attack image-level suite of \cref{tab:geom_full}
(robustness), and context-clustering rule-inference accuracy (structural
secrecy).

The first two axes vary monotonically in opposition across the full sweep,
with post-removal AUROC rising from $0.313$ to $0.974$ as image-attack
AUROC moves gently from $0.957$ to $0.884$, showing that
$\lambda_\mathrm{pc}$ acts as a smooth, predictable control knob on the
trade-off of Sec.~3.2 within a single architecture. Context-aware
rule inference remains near chance throughout ($0.511$ to $0.561$, vs.\
$0.986$ for KGW $h{=}0$ and $0.866$ for chunk-hash in
Tab.~1).

\paragraph{Why $\lambda_\mathrm{pc}{=}2.5$.}
The frontier exhibits a clear knee at $\lambda_\mathrm{pc}{=}2.5$. Raising
$\lambda_\mathrm{pc}$ from $1.0$ to $2.5$ gains $+0.064$ post-removal
AUROC for only $-0.020$ image-attack AUROC, with non-overlapping seed
ranges ($[0.869, 0.941]$ vs.\ $[0.954, 0.976]$). Increasing further to
$5.0$ adds just $+0.010$ post-removal AUROC (within the $\pm 0.018$ seed
std) while trading off an additional $2.0$\,pp on $5^\circ$ rotation
($0.837 \to 0.817$). Meanwhile, $\lambda_\mathrm{pc}{=}1.0$ provides a
convenient alternative operating point for deployments prioritizing
spatial perturbations ($0.907$ image-attack avg at $0.900$ post-removal
AUROC).

\paragraph{Operational vs.\ clustering secrecy.}
Notably, $\lambda_\mathrm{pc}{=}0$ yields $0.511$ context-clustering
accuracy yet drops to $0.313$ AUROC under global frequency counting:
without cross-position decorrelation, all steps converge toward a shared
global greenlist akin to $h{=}0$, leaving little context-specific
variation beyond the global list while exposing that global list directly
to marginal token counting. Evaluating both context-aware rule inference
and post-removal AUROC is therefore essential, and we report both
throughout the paper.

\begin{table*}[!t]
\centering
\footnotesize
\setlength{\tabcolsep}{5.5pt}
\caption{Monotone Pareto sweep over the cross-position decorrelation
weight $\lambda_\mathrm{pc}$ on LlamaGen-B ($\delta{=}4$; mean $\pm$ std
over three independent training seeds per setting under the shared
ablation recipe of \cref{sec:supp_ablation}). Increasing
$\lambda_\mathrm{pc}$ monotonically strengthens adaptive removal
resistance while smoothly trading off spatial-drift sensitivity; rule
inference stays near chance throughout.}
\label{tab:lpc_pareto}
\begin{tabular}{@{}lccccc@{}}
\toprule
$\lambda_\mathrm{pc}$ & Clean AUROC & \textbf{Removed} AUROC
  & Image-attack avg & Rot $5^\circ$ AUROC & Rule inf.\ acc. \\
\midrule
$0.0$           & $0.985 \pm 0.010$ & $0.313 \pm 0.030$        & $\best{0.957} \pm 0.001$ & $\best{0.966} \pm 0.001$ & $\best{0.511} \pm 0.002$ \\
$0.5$           & $0.989 \pm 0.006$ & $0.795 \pm 0.054$        & $0.923 \pm 0.007$        & $0.906 \pm 0.010$        & $0.529 \pm 0.005$ \\
$1.0$           & $0.992 \pm 0.003$ & $0.900 \pm 0.030$        & $0.907 \pm 0.003$        & $0.858 \pm 0.021$        & $0.529 \pm 0.006$ \\
$2.5$ (default) & $0.996 \pm 0.004$ & $0.965 \pm 0.009$        & $0.887 \pm 0.002$        & $0.837 \pm 0.003$        & $0.538 \pm 0.002$ \\
$5.0$           & $\best{0.998} \pm 0.002$ & $\best{0.974} \pm 0.018$ & $0.884 \pm 0.002$ & $0.817 \pm 0.003$        & $0.561 \pm 0.006$ \\
\bottomrule
\end{tabular}
\end{table*}

\subsection{Design space (three panels)}
\label{supp:design}

\begin{figure*}[!t]
\centering
\includegraphics[width=\textwidth]{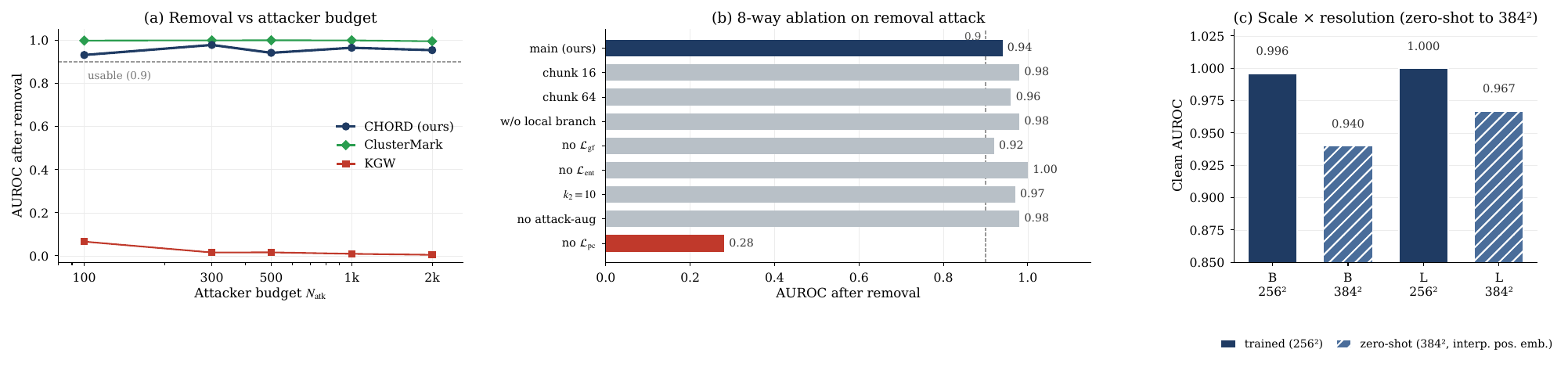}
\caption{\textbf{Design space of \method{}, three panels.}
\textbf{(a)}~Removal AUROC across attacker budgets
$N_\mathrm{atk}{\in}\{100,\dots,2000\}$ on LlamaGen-B ($\delta{=}4$):
KGW $h{=}0$ stays ${\leq}0.07$ at every budget while \method{} and CM hold
${>}0.9$ (Sec.~4.3).
\textbf{(b)}~Eight-way ablation on the same attack: only removing
\lpc{} (red bar, $0.28$) breaks the method; every other variant stays
within $0.06$ AUROC of the full model (Sec.~4.4).
\textbf{(c)}~Clean AUROC across generator sizes
($111$M\,/\,$343$M) and resolutions ($256^2\,/\,384^2$); $384^2$ bars
(hatched) reuse the same $T_\theta$ trained at $256^2$
(Sec.~4.4).}
\label{fig:design}
\end{figure*}

\cref{fig:design} collects in one place the three sweeps that bound the
design: how the removal attack behaves as the attacker's budget grows,
which of the eight ablated components carries removal resistance, and
whether the scheme survives a change of generator size and resolution.
Each panel summarizes a corresponding table in this supplement
(\cref{tab:natk_full,tab:ablation_removal,tab:scaling,tab:384_edits}).

\subsection{Scale and resolution transfer}
\label{supp:scaletab}

Retraining $T_\theta$ on $5\,000$ clean sequences from LlamaGen-L with the
exact same loss weights and schedule as LlamaGen-B, requiring no
per-model hyperparameter tuning, preserves both the edit and attack
robustness of Tab.~3 and the removal resistance of
\cref{tab:removal_full} ($0.943$ on B $\to$ $0.965$ on L
post-removal AUROC; \cref{tab:scaling}). Moreover, \method{} supports
zero-shot cross-resolution transfer across both clean and edited images:
applying the single $256^2$-trained $T_\theta$ directly to the $384^2$
($576$-token) LlamaGen-B and LlamaGen-L checkpoints via linear
positional-embedding interpolation achieves high post-edit
discriminability without any retraining (\cref{tab:384_edits}), reaching
$0.941$ mean aligned-edit AUROC on LlamaGen-L-$384$ and $0.869$ / $0.897$
AUROC under $5^\circ$ rotation on L-$384$ / B-$384$ (exceeding $256^2$
rotation AUROC as the $2.25\times$ longer sequence averages over $18$
chunks instead of $8$).

\begin{table}[!htb]
\centering
\footnotesize
\caption{Transfer of \method{} from LlamaGen-B ($111$M) to LlamaGen-L
($343$M) at $256^2$ ($\delta{=}4$) with identical $T_\theta$
hyperparameters, shown alongside ClusterMark (CM).}
\label{tab:scaling}
\setlength{\tabcolsep}{4.0pt}
\begin{tabular}{@{}lcccc@{}}
\toprule
Metric & \method{} B & \method{} L & CM B & CM L \\
\midrule
clean AUROC          & 0.996 & \best{1.000} & 1.000 & 1.000 \\
clean $\TPR@1\%$     & 0.975 & \best{1.000} & 1.000 & 1.000 \\
removed AUROC        & 0.943 & \best{0.965} & \best{1.000} & 0.995 \\
removed $\TPR@1\%$   & \best{0.850} & 0.773        & \best{0.997} & 0.970 \\
\bottomrule
\end{tabular}
\end{table}

\begin{table}[!htb]
\centering
\scriptsize
\setlength{\tabcolsep}{4.0pt}
\caption{\textbf{Zero-shot post-edit transfer to $384{\times}384$
($576$ tokens)} on LlamaGen-B and LlamaGen-L ($120$ paired images per
model, $\delta{=}4$), reusing the single $256^2$-trained LlamaGen-B
$T_\theta$ via linear positional-embedding interpolation without
retraining. Cells report AUROC/matched $\TPR@1\%$.}
\label{tab:384_edits}
\begin{tabular}{@{}l|cc@{}}
\toprule
Edit & LlamaGen-B ($384^2$) & LlamaGen-L ($384^2$) \\
\midrule
clean (no edit)             & 0.940/0.933 & 0.967/0.967 \\
\midrule
\multicolumn{3}{@{}l}{\itshape Aligned edits} \\
\quad JPEG $q25$            & 0.935/0.775 & 0.961/0.900 \\
\quad WebP $q25$            & 0.815/0.717 & 0.892/0.792 \\
\quad color jitter          & 0.896/0.850 & 0.898/0.617 \\
\quad brightness            & 0.902/0.875 & 0.913/0.858 \\
\quad saturation            & 0.939/0.933 & 0.967/0.958 \\
\quad noise $\sigma{=}0.05$ & 0.980/0.808 & 0.966/0.625 \\
\quad blur $r{=}2$          & 0.939/0.900 & 0.965/0.942 \\
\quad resize $50\%$         & 0.939/0.925 & 0.967/0.958 \\
\cmidrule(lr){1-3}
\quad \textit{mean (aligned)} & 0.918/0.848 & 0.941/0.831 \\
\midrule
\multicolumn{3}{@{}l}{\itshape Misaligning edits} \\
\quad rotate $5^\circ$      & 0.897/0.275 & 0.869/0.267 \\
\quad crop $5\%$            & 0.890/0.383 & 0.882/0.183 \\
\cmidrule(lr){1-3}
\quad \textit{mean (misaligning)} & 0.893/0.329 & 0.875/0.225 \\
\bottomrule
\end{tabular}
\end{table}

\subsection{Training-free continuous chunk control and cross-tokenizer verification}
\label{supp:rar}

\paragraph{Training-free continuous chunk baseline on LlamaGen-B.}
In Tab.~1, the exact \emph{chunk-hash} baseline hashes
the discrete token tuple of the preceding chunk. To further isolate the
contribution of our learned smooth network $T_\theta$ from the continuous
codebook representation itself on the \emph{same} LlamaGen-B generator, we
also evaluate a training-free continuous chunk baseline
(\emph{quantized-chunk}): it computes the exact same normalized mean
codebook embedding $\mathbf{c}_{k-1}\in\mathbb{R}^d$ (Eq.~(3),
$C{=}32$, $\gamma{=}0.25$), applies uniform scalar quantization
($\mathrm{round}(10\,\mathbf{c}_{k-1})$) to absorb minor embedding jitter,
and hashes the quantized vector with the secret key to draw a greenlist.
On $200$ LlamaGen-B images ($\delta{=}6$), \emph{quantized-chunk} reaches
only $0.950$ clean AUROC and $0.660$ clean $\TPR@1\%$ (versus $1.000$ and
$1.000$ for \method{} at $\delta{=}4$), because even slight VQ round-trip
shifts push continuous summaries across hard quantization bin boundaries;
moreover, under a $300$-query contextual codebook-cluster replacement
attack, its AUROC drops to $0.764$ ($\TPR@1\%=0.060$), whereas \method{}
retains $0.98$ at the same $300$-query budget (\cref{tab:natk_full}).
Thus, pooling continuous codebook embeddings
alone is insufficient without a learned smooth, decorrelated mapping
$T_\theta$.

\paragraph{Cross-tokenizer verification on RAR-XL.}
To verify that neither our continuous chunk representation nor the full
learned $T_\theta$ pipeline is tied to LlamaGen's VQ-16 codebook
($V{=}16\,384$, $8$-dim embeddings), we evaluate both the training-free
\emph{quantized-chunk} prototype and a dedicated $T_\theta$ trained with
the full \method{} objective on RAR-XL~\cite{yu2025rar}, which uses the
maskgit-vqgan tokenizer~\cite{chang2022maskgit} with $V{=}1\,024$ codes and
$256$-dim embeddings ($N{=}256$ tokens), representing a $16\times$ smaller
vocabulary and a $32\times$ higher embedding dimension. Because both
$\mathbf{c}_{k-1}$ and $T_\theta$'s input projection operate on normalized
codebook vectors, porting requires zero architectural changes ($5\,000$
training steps on clean RAR-XL token sequences with the default loss
weights of Sec.~3.3). Across $300$ class-conditional RAR-XL
images ($C{=}32$, $\gamma{=}0.25$; \cref{tab:rar}), the training-free
prototype ($\delta{=}5$) already reaches $0.900$ clean AUROC ($0.727$
$\TPR@10\%$) and $0.734$ / $0.642$ under $5\%$ cropping and $5^\circ$
rotation. Training the dedicated $T_\theta$ ($\delta{=}4$) substantially
improves every condition even at a lower logit bias: clean AUROC rises to
$0.959$ ($0.883$ $\TPR@10\%$), aligned edits reach $0.922$ mean AUROC across
all eight aligned transforms of Tab.~1 (including
$+0.126$ AUROC on JPEG $q25$, $+0.147$ on Gaussian noise, $0.955$ on
saturation, $0.952$ on brightness, $0.931$ on color jitter, and $0.867$ on
WebP $q25$), and misaligning geometric edits jump by $+0.203$ AUROC on
$5\%$ cropping ($0.937$, $\TPR@10\%=0.820$) and $+0.257$ AUROC on
$5^\circ$ rotation ($0.899$, $\TPR@10\%=0.680$), while $64$-cluster
context-aware rule inference ($N_\mathrm{atk}{=}500$) stays near chance at
$0.526$ balanced accuracy. Together, these results confirm that \method{}
generalizes cleanly across markedly different tokenizer vocabularies,
embedding dimensions, and AR generator families.

\begin{table}[!htb]
\centering
\footnotesize
\setlength{\tabcolsep}{3.8pt}
\caption{Cross-tokenizer post-edit detection on RAR-XL ($V{=}1\,024$,
$256$-dim maskgit-vqgan codebook, $300$ images, $C{=}32$, $\gamma{=}0.25$),
comparing the training-free quantized-chunk prototype ($\delta{=}5$)
against full learned \method{} ($\delta{=}4$, zero architectural changes).
Full \method{} improves every edit even at lower bias $\delta$, gaining
$+0.203$ and $+0.257$ AUROC on cropping and rotation while holding
$64$-cluster rule inference at $0.526$ balanced accuracy.}
\label{tab:rar}
\begin{tabular}{@{}l cc cc@{}}
\toprule
 & \multicolumn{2}{c}{Quantized-chunk ($\delta{=}5$)}
 & \multicolumn{2}{c}{\oc{\textbf{\method{} ($\delta{=}4$)}}} \\
\cmidrule(lr){2-3}\cmidrule(lr){4-5}
Edit & AUROC & $\TPR@10\%$ & \oc{AUROC} & \oc{$\TPR@10\%$} \\
\midrule
clean                   & 0.900 & 0.727 & \oc{\best{0.959}} & \oc{\best{0.883}} \\
\midrule
resize $50\%$           & 0.904 & 0.713 & \oc{\best{0.948}} & \oc{\best{0.860}} \\
blur $r{=}2$            & 0.861 & 0.617 & \oc{\best{0.947}} & \oc{\best{0.867}} \\
JPEG $q25$              & 0.760 & 0.440 & \oc{\best{0.886}} & \oc{\best{0.570}} \\
noise $\sigma{=}0.05$   & 0.747 & 0.413 & \oc{\best{0.894}} & \oc{\best{0.633}} \\
crop $5\%$              & 0.734 & 0.410 & \oc{\best{0.937}} & \oc{\best{0.820}} \\
rotate $5^\circ$        & 0.642 & 0.237 & \oc{\best{0.899}} & \oc{\best{0.680}} \\
\bottomrule
\end{tabular}
\end{table}

\subsection{Fidelity (FID and IS)}
\label{sec:supp_fid}

\cref{tab:fid} reports Fr\'echet Inception Distance (FID) and Inception
Score (IS) on $5\,000$ LlamaGen-L samples relative to a size-matched
clean baseline. All three token-level watermarks lie within a narrow
band ($1.06$ to $1.12\times$ clean FID) and preserve Inception Score
around $69$ to $71$, confirming that \method{}'s position-varying bias
maintains generation quality comparable to static-partition baselines
(matched-seed visual examples appear in Fig.~4).

\begin{table}[!htb]
\centering
\footnotesize
\caption{Generation fidelity on LlamaGen-L ($5\,000$ samples). $\Delta$
denotes FID relative to the size-matched unwatermarked reference.}
\label{tab:fid}
\setlength{\tabcolsep}{4pt}
\begin{tabular}{@{}lccc@{}}
\toprule
Method & FID~$\downarrow$ & $\Delta$ & IS~$\uparrow$ \\
\midrule
clean (ref.)             & 8.86 & $1.00\times$ & 69.9 \\
KGW $h{=}0$              & 9.36 & $1.06\times$ & 70.2 \\
CM                       & 9.78 & $1.10\times$ & 70.7 \\
\method{} ($\delta{=}4$) & 9.90 & $1.12\times$ & 69.0 \\
\bottomrule
\end{tabular}
\end{table}

\subsection{Runtime analysis}
\label{supp:runtime}

\begin{table}[!htb]
\centering\footnotesize
\setlength{\tabcolsep}{3.5pt}
\caption{\textbf{Inference runtime on LlamaGen-B ($256^2$, A100).}
Generation at batch~$8$, verification at batch~$4$, per image.
Verification is dominated by the shared VQ encoder ($T_\theta$:
${\approx}2.5$\,ms/img); generation-side cost is one-time at creation
and fixed across generator scales.}
\label{tab:runtime}
\begin{tabular}{@{}lrr@{}}
\toprule
Method & Gen.\,(ms/img) & Verif.\,(ms/img) \\
\midrule
Clean (no watermark)                & $381.3$ & ---   \\
KGW~\cite{kirchenbauer2023kgw}      & $381.6$ & $5.4$ \\
IM~\cite{tong2025indexmark}         & $382.5$ & $4.5$ \\
CM~\cite{lukovnikov2026clustermark} & $381.5$ & $5.5$ \\
\method{} (ours)                    & $525.5$ & $8.0$ \\
\bottomrule
\end{tabular}
\end{table}

\cref{tab:runtime} benchmarks generation and verification latency on a
single NVIDIA A100 GPU. Verification---the latency-critical operation in
provenance checking, run on every queried image---is dominated by the
shared VQ re-encoding pass required by all token-level watermarks;
\method{} adds less than $3$\,ms per image on top of it ($8.0$\,ms
total). On the generation side, the lightweight $T_\theta$ MLP adds
$0.56$\,ms per token ($144$\,ms per image), a one-time provider-side
cost at creation: the compute footprint of context-conditioned signaling,
whereas static-hash baselines are near-free precisely because their
fixed table-lookup rule is also what makes them recoverable and removable
(Sec.~4.3). Because $T_\theta$ depends only on the codebook size $V$ and
not on the generator size, this absolute cost stays constant as the AR
generator scales, an increasingly small relative fraction on larger
models.

\end{document}